%% file: main.tex
\documentclass[conference]{IEEEtran}
\IEEEoverridecommandlockouts

\input{config}

\def\BibTeX{{
    \rm B
    \kern-.05em{\sc i\kern-.025em b}
    \kern-.08em T
    \kern-.1667em\lower.7ex\hbox{E}
    \kern-.125emX}
}

\definecolor{mygray}{gray}{0.95}
\definecolor{mydarkgray}{gray}{0.85}

\begin{document}

\title{
A Continual Offline Reinforcement Learning
Benchmark for Navigation Tasks
}   


    \author{
        \IEEEauthorblockN{
            Anthony Kobanda\textsuperscript{1,2},
            Odalric-Ambrym Maillard\textsuperscript{1},
            Rémy Portelas\textsuperscript{2}
        }
        \IEEEauthorblockA{
            \textsuperscript{1} \textit{Inria, Univ. Lille, CNRS, Centrale Lille, UMR 9198-CRIStAL, F-59000 Lille, France}
        }
        \IEEEauthorblockA{
            \textsuperscript{2} \textit{Ubisoft La Forge, Bordeaux, France}
        }
        \{anthony.kobanda,remy.portelas\}@ubisoft.com , odalric.maillard@inria.fr
    }

    \maketitle


    \input{1-FRONT/1-abstract}

    \input{1-FRONT/2-introduction}

    \input{1-FRONT/3-related_work}

    \input{1-FRONT/4-preliminaries}


    \input{2-MAIN/1-environments}

    \input{2-MAIN/2-baselines}

    \input{2-MAIN/3-benchmark}

    \input{2-MAIN/4-discussion}

    
    \newpage
    \bibliographystyle{ieeetr}
    \bibliography{bib}

\end{document}

%% file: config.tex
\usepackage[utf8]{inputenc}

\usepackage[T1]{fontenc}

\usepackage[caption=false]{subfig}

\usepackage{enumitem}

\usepackage{amsmath,amsfonts,amssymb,bbm,mathtools}

\usepackage{graphicx}
\usepackage{dsfont}
\usepackage{float}

\usepackage{hyperref}
\usepackage{cite}

\usepackage{url}

\usepackage{array}
\usepackage{booktabs}
\usepackage{caption}
\usepackage{floatflt}
\usepackage{multirow}
\usepackage{makecell}
\usepackage{subfig}
\usepackage{tabularray}
\usepackage{wrapfig}

\usepackage{nicefrac}

\usepackage{microtype}

\usepackage[dvipsnames]{xcolor}

\usepackage{algorithm}
\usepackage{algorithmic}
\usepackage{cite}
\usepackage{textcomp}

%% file: 1-FRONT/1-abstract.tex
\begin{abstract}
Autonomous agents operating in domains such as robotics or video game simulations must adapt to changing tasks without forgetting about the previous ones. This process called Continual Reinforcement Learning poses non-trivial difficulties, from preventing catastrophic forgetting to ensuring the scalability of the approaches considered.
Building on recent advances, we introduce a benchmark providing a suite of video-game navigation scenarios, thus filling a gap in the literature and capturing key challenges : catastrophic forgetting, task adaptation, and memory efficiency. 
We define a set of various tasks and datasets, evaluation protocols, and metrics to assess the performance of algorithms, including state-of-the-art baselines.
Our benchmark is designed not only to foster reproducible research and to accelerate progress in continual reinforcement learning for gaming, but also to provide a reproducible framework for production pipelines -- helping practitioners to identify and to apply effective approaches.\\\\
{\color{PineGreen}\small\url{https://sites.google.com/view/continual-nav-bench}}\\
\end{abstract}

\vspace{-0.5em}

%% file: 1-FRONT/2-introduction.tex
\section{Introduction}
\label{sec:introduction}

The human learning process is inherently cumulative : we\linebreak 
constantly master skills through new experiences without discarding what we already know, instead we leverage previously acquired knowledge. In contrast, most classical \textbf{Reinforcement Learning (RL)} \cite{rl,deeprl} methods depend on extensive interactions and are prone to catastrophic forgetting when faced with a series of different challenges. In this context,\linebreak
\textbf{Continual Reinforcement Learning (CRL)} \cite{crl_survey,crl_survey2} aim to bridge this gap by enabling agents to learn incrementally, thus emulating the adaptive, lifelong learning process of humans. This incremental paradigm not only helps mitigate forgetting but also facilitates the transfer of learned skills.\\

An interesting subset of tasks within the CRL framework is \textbf{Goal-Conditioned RL (GCRL)} \cite{gcbc,gcbcsurvey} allowing policies to be conditioned to specific target states, which is particularly valuable in navigation tasks. Modern video game environments, with their evolving layouts require efficiently learning and adapting bots. Moreover, in production pipelines, rigorous evaluation of these methods is critical to ensure that algorithms are effective and computationally efficient. These attributes make GCRL particularly promising for applications that demand adaptive and robust decision-making strategies.

\begin{figure}[t!]
    \centering
    \begin{minipage}[h!]{0.32\linewidth}
        \centering
        \includegraphics[width=1\linewidth]{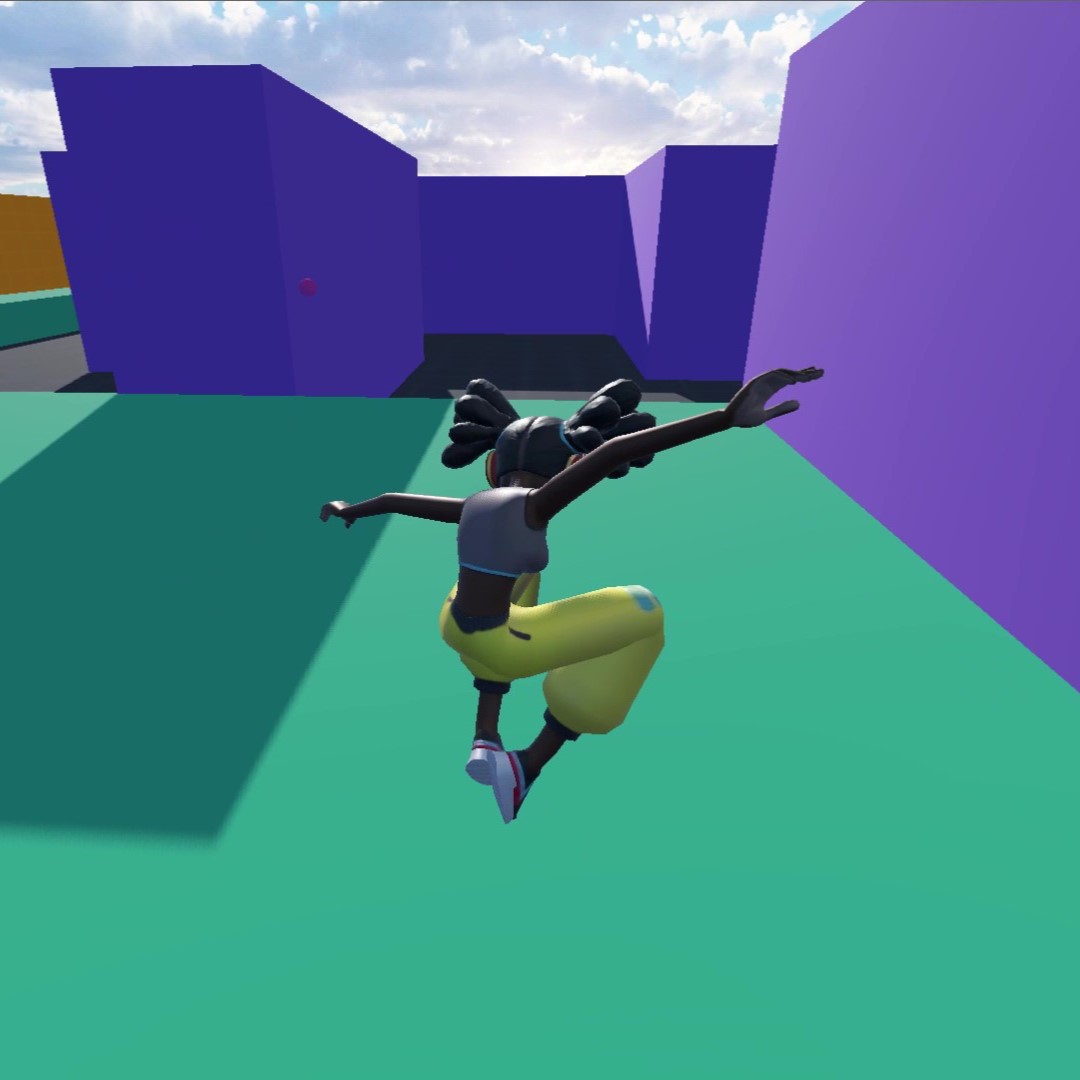}
        \vspace{0.35em}
        \subfloat{\footnotesize(A) A human playing.}
        \label{fig:intro:sub_A}
    \end{minipage}
    \hfill
    \begin{minipage}[h!]{0.32\linewidth}
        \centering
        \includegraphics[width=1\linewidth]{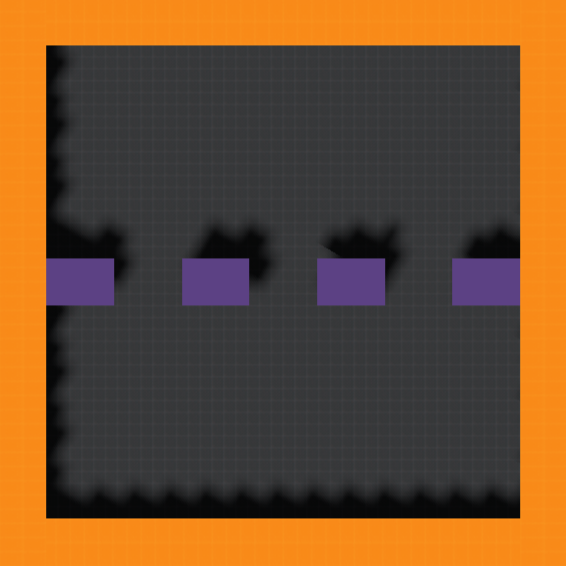}
        \vspace{0.35em}
        \subfloat{\footnotesize(B)$\ 20\ m\times20\ m$ map.}
        \label{fig:intro:sub_B}
    \end{minipage}
    \hfill
    \begin{minipage}[h!]{0.32\linewidth}
        \centering
        \includegraphics[width=1\linewidth]{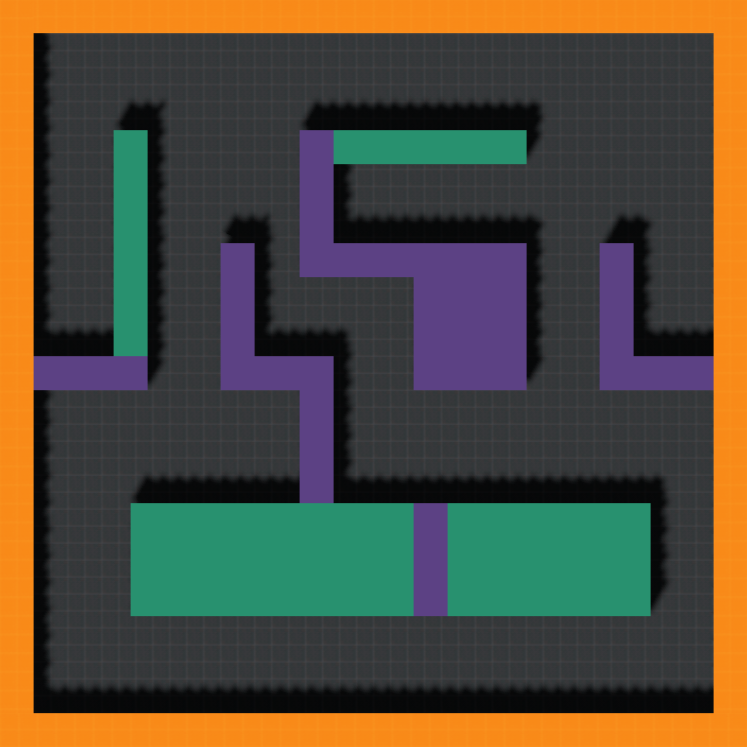}
        \vspace{0.35em}
        \subfloat{\footnotesize(B)$\ 60\ m\times60\ m$ map.}
        \label{fig:intro:sub_C}
    \end{minipage}
\vspace{0.15em}
\caption{\small
\textbf{Visualization of an environment and a human playing :}\linebreak
\textbf{(a) Human Player Interaction.} A third-person perspective of a human playing within a large maze ;
\textbf{(b) Overview of a Small Map.} A top-down view of a small maze, showcasing the simple layout ;\linebreak
\textbf{(c) Overview of a Large Map.} A top-down view of a larger maze, highlighting a complex layout requiring better planning and actions.
}
\label{fig:intro}
\end{figure}

\textbf{Offline RL (ORL)} \cite{offrl,bdt} offers a compelling framework by leveraging collections of pre-recorded gameplay or simulation data, reducing the cost and risks of live data acquisition. 
However, despite ORL’s progress, standardized
benchmarks for CRL in video game–inspired navigation tasks
remain scarce. Current benchmarks \cite{towardbench,trackbench} rarely address the
flexibility and memory requirements of production pipelines.\\

\vspace{-0.75em}

In this work, we introduce {\color{PineGreen}\textbf{Continual NavBench}} : a novel benchmark for Continual Offline Reinforcement Learning focused on navigation in virtual game worlds. Continual NavBench bridges research and production needs by offering\linebreak
a suite of scenarios -- including Godot engine \cite{godot} tailor made\linebreak
environments with human-generated data -- for both research and real-world applications. Our benchmark includes :

\vspace{0.33em}

\begin{itemize}[leftmargin=*] 
    \item {\color{PineGreen}\underline{\textbf{Standardized Offline Datasets :}}} Sourced from 10 hours of human gameplay (around 3000 episodes) in diverse Godot mazes, to capture different navigation strategies. 
    \vspace{0.6em}
    \item {\color{PineGreen}\underline{\textbf{Evaluation Protocols and Metrics :}}} We consider standard continual metrics -- the \textit{Performance}, the \textit{Backward Transfer}, the \textit{Forward Transfer}, the \textit{Relative Model Size}, the \textit{Training and Inference Costs} -- to assess a model performance and efficiency, ensuring relevance for video game production.
    \vspace{0.6em}
    \item {\color{PineGreen}\underline{\textbf{Reproducible Benchmarking Tools :}}} Our open - source framework provides code, datasets, and evaluation protocols to ensure reproducible experiments and facilitate comparisons of methods for streamlined  integration into research pipelines. 
\end{itemize}

%% file: 1-FRONT/3-related_work.tex
\section{Related Work}
\label{sec:related-work}

\textbf{Reinforcement Learning (RL)} has achieved notable successes in domains such as robotics and video games \cite{rl,deeprl}. Benchmarks, such as Atari 2600 \cite{atari}, Procgen \cite{procgen},
and VizDoom \cite{vizdoom}, evaluate RL agents in dynamic and rich environments. However, classical methods rely suffers from catastrophic forgetting when agents are faced with sequential tasks. \textbf{Continual Reinforcement Learning (CRL)} \cite{crl_survey,crl_metrics} address these issues by enabling agents to learn incrementally, better approximating the lifelong learning exhibited by humans.\\

In particular, \textbf{Goal-Conditioned RL (GCRL)} \cite{gcbc,gcbcsurvey} shows promise for navigation tasks by conditioning policies on desired goal states. Despite these advances, most CRL research has focused on the online setting. In contrast, \textbf{Offline RL} \cite{offrl,corl-dualreplay} leverages fixed datasets of pre-recorded gameplay or simulation traces, bypassing the risks and costs of live data collection. Nevertheless, standardized benchmarks for Offline CRL in video game–inspired navigation tasks remain scarce. Existing benchmarks, such as Continual World \cite{conworld} and CORA \cite{cora}, primarily target online learning and rarely address production requirements and constraints like inference speed, memory efficiency, and model scalability.\\

CRL methods address the challenge of sequential task learning through various strategies : \textbf{Replay-Based} methods mitigate forgetting by storing past experiences and replaying them \cite{replay_crl, mbased_crl}, however they can be impractical due to significant storage requirements and potential privacy concerns, especially in industrial applications ; \textbf{Regularization} techniques such as Elastic Weight Consolidation (EWC) \cite{ewc}, or simply\linebreak
L2 regularization \cite{l2reg} introduce constraints on parameter updates, but they may struggle with highly diverse tasks.\linebreak
\textbf{Architectural} strategies, such as Progressive Neural Networks (PNNs) \cite{pnn} or \cite{hispo}, modify a networks architecture to accommodate new tasks, nevertheless they can become memory-intensive and may lack scalability.\\

%% file: 1-FRONT/4-preliminaries.tex
\section{Preliminaries}
\label{sec:preliminaries}

In this section, we present the theoretical background of CRL.
These foundational concepts provide the basis for understanding and comparing the performance of different approaches.

\subsection{Reinforcement Learning}

We consider a Markov Decision Process (MDP) framework $\mathcal{M} = \big(\ \mathcal{S},\ \mathcal{A},\ \mathcal{P}_{\mathcal{S}},\ {\mathcal{P}_\mathcal{S}}^{(0)},\ \mathcal{R},\ \gamma\ \big)$,
which provides a formal framework for RL,
 where 
$\mathcal{S}$ is a state space, 
$\mathcal{A}$ an action space, 
$\mathcal{P}_{\mathcal{S}} : \mathcal{S} \times \mathcal{A} \rightarrow \Delta(\mathcal{S})$ a transition function,
${\mathcal{P}_{\mathcal{S}}}^{(0)} \in \Delta(S)$ an initial distribution over the states, 
$\mathcal{R} : \mathcal{S} \times \mathcal{A} \times \mathcal{S} \rightarrow \mathbb{R}$ a deterministic reward function, and $\gamma\in\ ]0,1 ]$ a discount factor.
An agent’s behavior follows a policy $\pi_\theta : \mathcal{S} \rightarrow \Delta(\mathcal{A})$, parameterized by $\theta\in\Theta$.
The objective is to learn a optimal parameters $\theta^*_\mathcal{M}$, or a set of such parameters, maximizing the expected cumulative reward $J_{\mathcal{M}}(\theta)$ or the success rate $\sigma_{\mathcal{M}}(\theta)$.

\subsection{Offline Goal-Conditioned Reinforcement Learning}

We extend the MDP to include a goal space $\mathcal{G}$, an initial state-goal distribution ${\mathcal{P}_{\mathcal{S},\mathcal{G}}}^{(0)}$, a function mapping each state to the goal it represents $\phi:\mathcal{S}\rightarrow\mathcal{G}$, and $d:\mathcal{G}\times\mathcal{G}\rightarrow\mathbb{R}^+$ a distance metric on $\mathcal{G}$. Then, the policy $\pi_\theta:\mathcal{S}\times\mathcal{G}\rightarrow\Delta(\mathcal{A})$ and the reward function $\mathcal{R} : \mathcal{S} \times \mathcal{A} \times \mathcal{S} \times \mathcal{G} \rightarrow \mathbb{R}$ now depend on a goal $g\in\mathcal{G}$. We consider sparse rewards allocated when agents reach a set goal : $\mathcal{R}(s_t,a_t,s_{t+1},g) = \mathbbm{1}\big(\ d(\phi(s_{t+1}),g)\ \scalebox{0.75}{$\leq$}\ \epsilon\ \big)$, and given a pre-collected dataset $\mathcal{D}=\big\{ (s,a,r,s',g)  \big\}$ \cite{offrl,rnddagger},
the policy is then optimized to reach specified goals.

\subsection{Continual Reinforcement Learning}

In CRL, an agent follows a sequence of tasks, or stream, $\mathcal{T}=\big(T_1,...,T_N\big)$, with $T_k=(\mathcal{M}_k,\mathcal{D}_k)$. 
We note as $\theta_k$ the parameters of a policy after learning on the $k$-th task.
As the agent learns new skills, it must preserve (to prevent forgetting) and enhance (to encourage backward transfer) its performance on tasks already learned, while ideally having a relatively low number of parameters.
To quantitatively compare CRL methods, we adopt standard metrics commonly used in the literature \cite{crl_metrics,crl_metrics2} :
{\small\textbf{Performance (PER) :} $\frac{1}{N}\sum_{k=1}^{N}\sigma_{\mathcal{M}_k}(\theta_N)$}; 
{\small \textbf{Backward Transfer (BWT) :} $\frac{1}{N}\sum_{k=1}^{N}\Big( \sigma_{\mathcal{M}_k}(\theta_N) - \sigma_{\mathcal{M}_k}(\theta_k) \Big)$};
{\small \textbf{Forward Transfer (FWT) :} $\frac{1}{N}\sum_{k=1}^{N}\Big( \sigma_{\mathcal{M}_k}(\theta_{k}) - \sigma_{\mathcal{M}_k}(\Tilde{\theta}_k) \Big)$} ;
{\small\textbf{Relative Model Size (MEM) :} $\frac{|\theta_N|}{|\theta_{\text{ref}}|}$};
\vspace{0.33em}
{\small\textbf{Model Inference Cost (INF)}};
{\small\textbf{Model Training Cost (TRN)}}.
{\small\textbf{PER}} measures the success rate across all tasks.
{\small\textbf{BWT}} indicates how learning a new task affects previous ones, while {\small\textbf{FWT}} measures the transfer of knowledge, using $\Tilde{\theta}_k$ as randomly initialized parameters.
{\small\textbf{MEM}} compares the memory load of the model to a reference one associated to parameters $\theta_{\text{ref}}$.
{\small\textbf{INF}} is the duration required for policy inference across tasks, and {\small\textbf{TRN}} is the overall training duration.

%% file: 2-MAIN/1-environments.tex
\section{Environments, Tasks \& Datasets}
\label{sec:envs}

\subsection{Navigation Environments \& Mazes}

\setlength{\skip\footins}{0.1pt}

We propose video game–inspired navigation environments, which offer easy to use 3D maze configurations via our provided open-source code\footnote{\href{https://github.com/anosubcog9438/continual-nav-bench}{GitHub Repository Link.}}. Our benchmark comprises two families of mazes designed to evaluate CRL agents :
{\textbf{SimpleTown}, a collection of $8$ relatively simple mazes ($20\ m\times20\ m$), and \textbf{AmazeVille}, a more complex set of $8$ mazes ($60\ m\times60\ m$).\linebreak
\textit{Table \ref{fig:envs:godot-act-features}} and \textit{Table \ref{fig:envs:godot-obs-features}} display the available outputs and inputs.

\input{2-MAIN/TABLES/godot-act-features}

\input{2-MAIN/FIGURES/maps}

\input{2-MAIN/TABLES/godot-obS-features}

\subsection{Task Streams}

Our benchmark defines task streams with different maze configurations. We distinguish two types of streams :\\

\textbf{Random Streams} – In AmazeVille we have :\\
\textit{Stream AR1 :} {\small
\texttt{A-LOOX} $\rightarrow$ \texttt{A-HXOX} $\rightarrow$ \texttt{A-LXOX} $\rightarrow$ \texttt{A-HXOX} ;
}\\
\textit{Stream AR2 :} {\small
\texttt{A-HXOO} $\rightarrow$ \texttt{A-HOOX} $\rightarrow$ \texttt{A-LOOX} $\rightarrow$ \texttt{A-LXOO}.
}\\

\textbf{Topological Streams} – Designed with common changes in maze structure. We have in both environments :\\
\textit{Stream AT1 :} {\small
\texttt{A-HOOX} $\rightarrow$ \texttt{A-HXOX} $\rightarrow$ \texttt{A-HXOX} $\rightarrow$ \texttt{A-HOOX} ;
}\\
\textit{Stream AT2 :} {\small
\texttt{A-LOOO} $\rightarrow$ \texttt{A-LOOO} $\rightarrow$ \texttt{A-LXOX} $\rightarrow$ \texttt{A-LXOO} ;
}\\
\textit{Stream ST1 :} {\small
\texttt{S-BASE} $\rightarrow$ \texttt{S-OXO} $\rightarrow$ \texttt{S-BASE} $\rightarrow$ \texttt{S-OOX} ;
}\\
\textit{Stream ST2 :} {\small
\texttt{S-BASE} $\rightarrow$ \texttt{S-OXX} $\rightarrow$ \texttt{S-XOO} $\rightarrow$ \texttt{S-OXX}.
}\\

Notably, some tasks reoccur within a stream, offering an opportunity to evaluate whether an algorithm can recognize and reuse previously learned strategies without increasing model size or compromising performance. We encourage future users to tailor task streams to address specific research or production goals. Additionally, these environments are well-suited for developing and benchmarking GCRL algorithms.

\subsection{Human Generated Datasets}

We gathered 2800 trajectories over 10 hours of human plays, with 250 episodes per maze in SimpleTown and 100 episodes in AmazeVille. These datasets capture strategies that are valuable for training bots to exhibit human-like behavior.\\

\begin{figure}[H]
    \vspace{-2.125em}
    \centering
    \begin{minipage}[h!]{\linewidth}
        \centering
        \includegraphics[width=\linewidth]{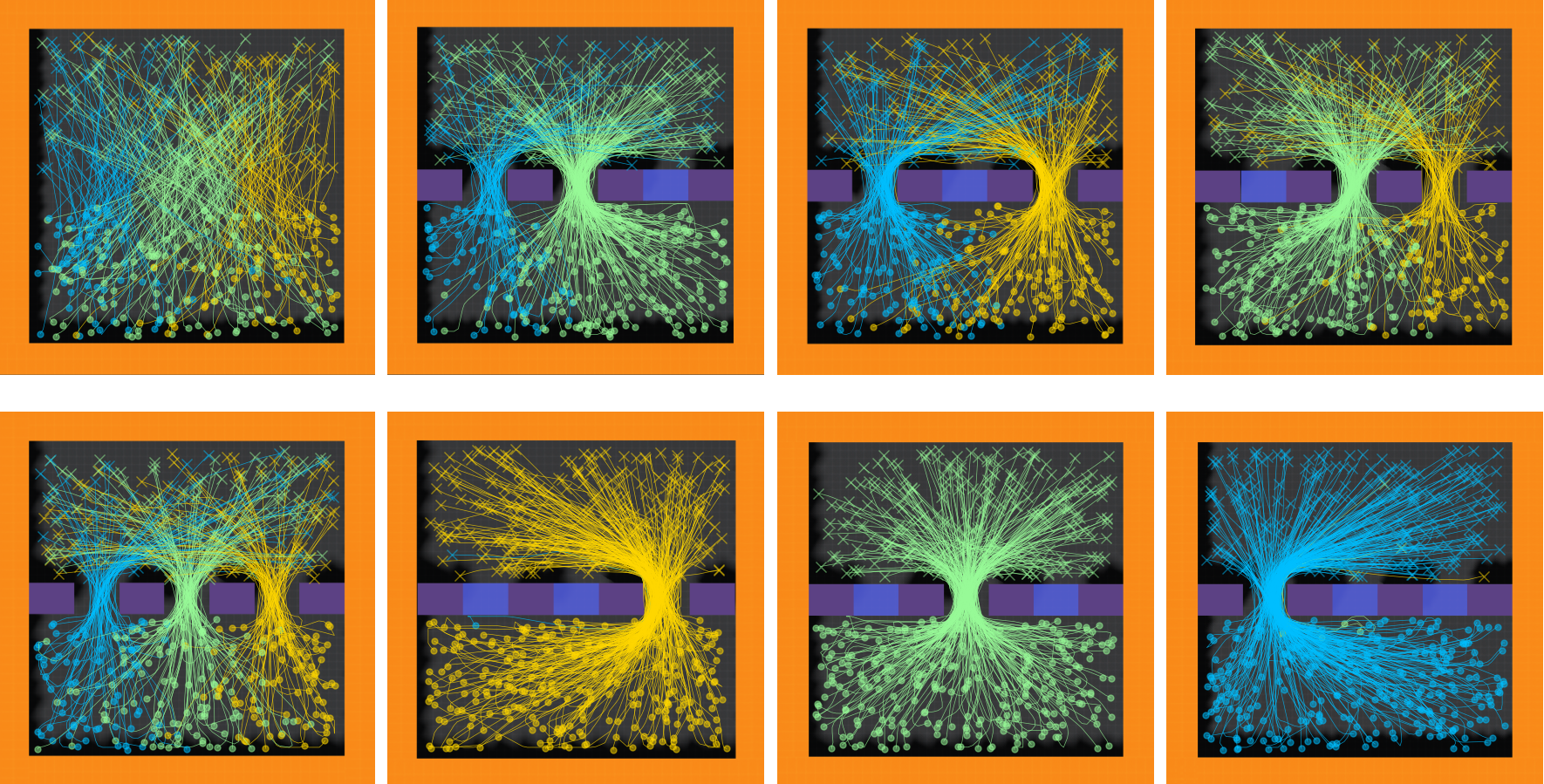}
        \subfloat{\footnotesize(A) Trajectories generated within SimpleTown mazes.}
        \label{fig:envs:maps_A}
        \vspace{0.75em}
    \end{minipage}
    \hfill
    \begin{minipage}[h!]{\linewidth}
        \centering
        \includegraphics[width=\linewidth]{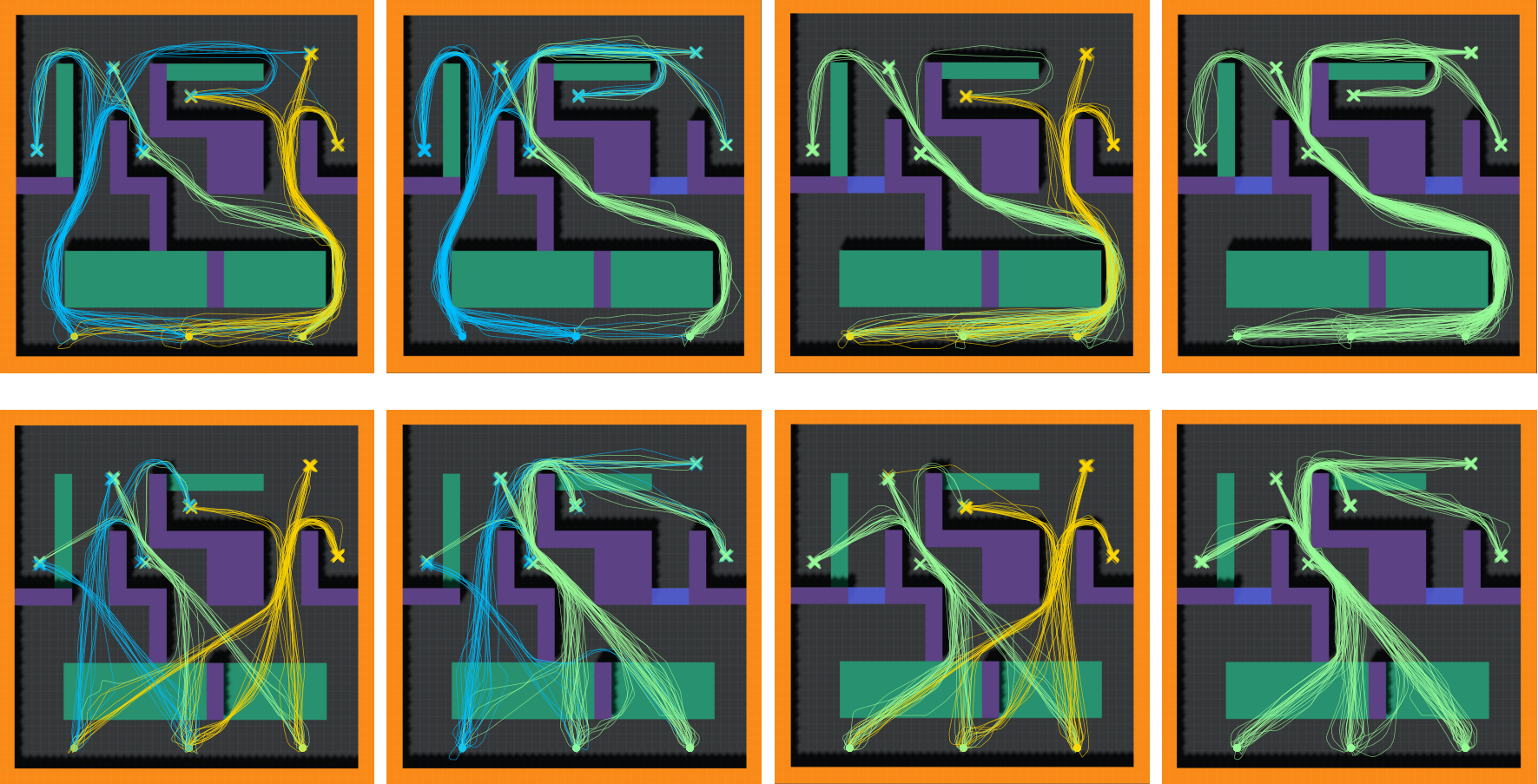}
        \subfloat{\footnotesize(B) Trajectories generated within AmazeVille mazes.}
        \label{fig:envs:maps_B}
    \end{minipage}
\caption{\small
\textbf{Visualization of the generated trajectories}
}
\label{fig:maps}
\end{figure}
\vspace{-1.33em}
To further support the research community, we open source the data alongside our benchmark framework.
This ensures that other practitioners can readily access, reproduce, and build upon our work, fostering collaborative advancements in CRL, and more generally GCRL, for video game bots development.

%% file: 2-MAIN/TABLES/godot-act-features.tex
\begin{table}[h]

    \vspace{0.15em}
    \scriptsize
    \centering
    
    \begin{tblr}{
        colspec = |c||c|c|c|,
        row{1} = {SeaGreen},
    }
    
        \hline
        \textbf{Action} & \textbf{Dimensions} & \textbf{Type} & \textbf{Human Controls / Keys} \\
        \hline
        \hline
        
        \ \ Move Forwards\ \  & $1$ &\  \texttt{bool}\  & Z \\
        \ \ Move Backwards\ \  & $1$ &\  \texttt{bool}\  & S \\
        \ \ Move Left\ \  & $1$ &\  \texttt{bool}\  & Q \\
        \ \ Move Right\ \  & $1$ &\  \texttt{bool}\  & D \\
        \ \ Jump\ \  & $1$ &\  \texttt{bool}\  & Space \\
        \ \ Turn\ \  & $1$ &\  \texttt{float}\  & Mouse \\

        \hline
        
    \end{tblr}
    \vspace{0.15em}
    \caption{
    \textbf{Available Action Keys.} 
    This table details the control commands available to the agents in our environments. 
    Actions include directional movement, running, jumping, and turning.
    The \textit{Keys} indicate the human input used during data collection.
    }
    \label{fig:envs:godot-act-features}

\end{table}

%% file: 2-MAIN/FIGURES/maps.tex
\begin{figure*}[t!]

    \centering
    \begin{minipage}[h!]{0.11\linewidth}
        \centering
        \includegraphics[width=1\linewidth]{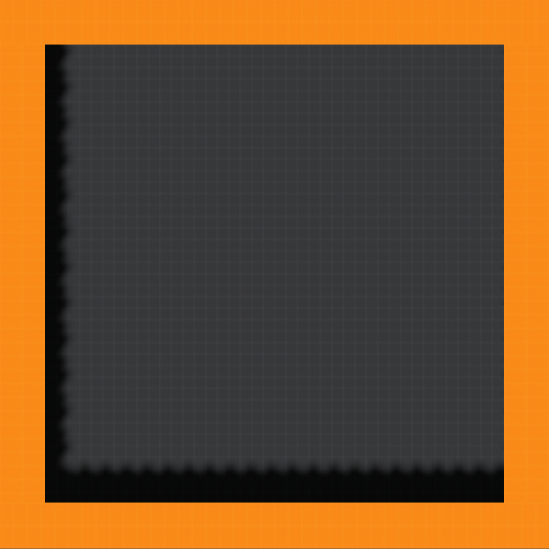}
        \subfloat{\footnotesize{S - BASE.}}
        \label{fig:envs:sbase}
        \vspace{1em}
    \end{minipage}
    \hfill
    \begin{minipage}[h!]{0.11\linewidth}
        \centering
        \includegraphics[width=1\linewidth]{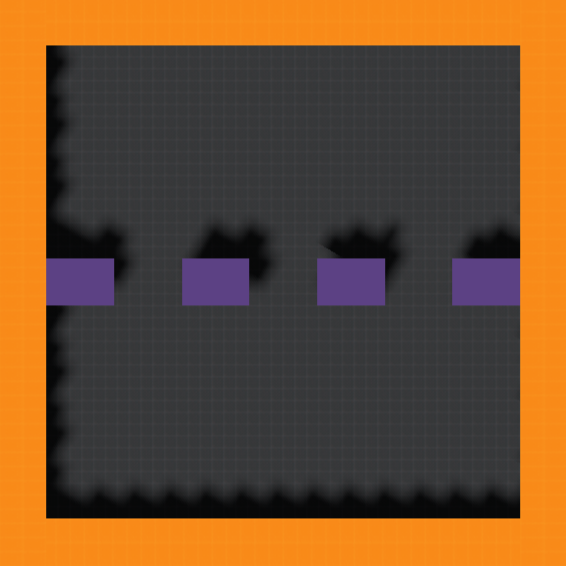}
        \subfloat{\footnotesize{S - OOO.}}
        \label{fig:envs:sooo}
        \vspace{1em}
    \end{minipage}
    \hfill
    \begin{minipage}[h!]{0.11\linewidth}
        \centering
        \includegraphics[width=1\linewidth]{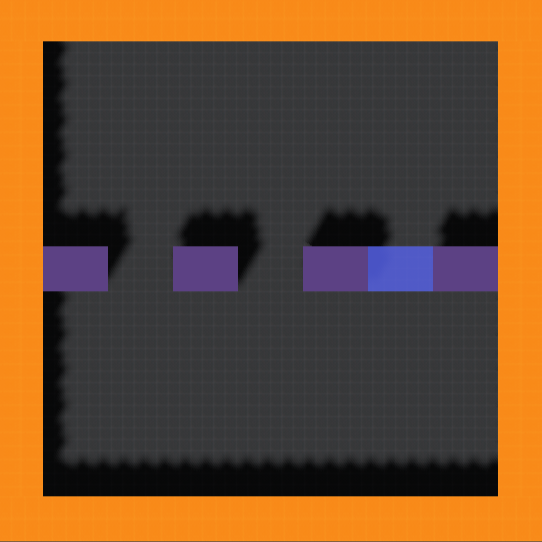}
        \subfloat{\footnotesize{S - OOX.}}
        \label{fig:envs:soox}
        \vspace{1em}
    \end{minipage}
    \hfill
    \begin{minipage}[h!]{0.11\linewidth}
        \centering
        \includegraphics[width=1\linewidth]{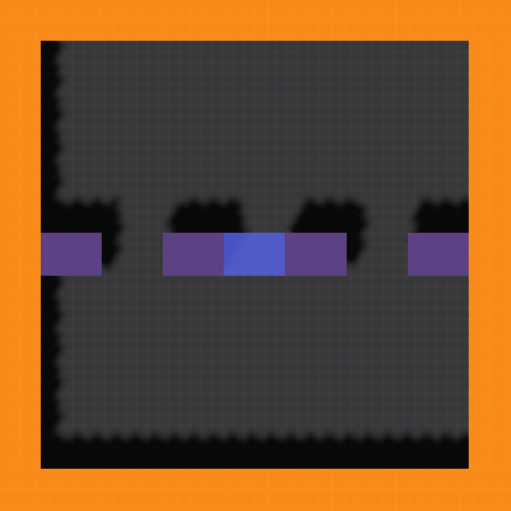}
        \subfloat{\footnotesize{S - OXO.}}
        \label{fig:envs:soxo}
        \vspace{1em}
    \end{minipage}
    \hfill
    \begin{minipage}[h!]{0.11\linewidth}
        \centering
        \includegraphics[width=1\linewidth]{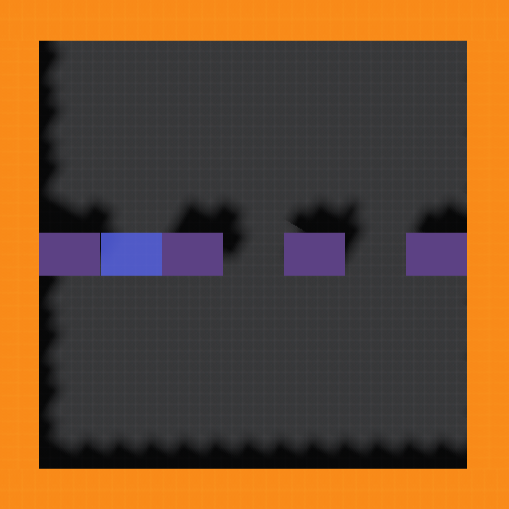}
        \subfloat{\footnotesize{S - XOO.}}
        \label{fig:envs:sxoo}
        \vspace{1em}
    \end{minipage}
    \hfill
    \begin{minipage}[h!]{0.11\linewidth}
        \centering
        \includegraphics[width=1\linewidth]{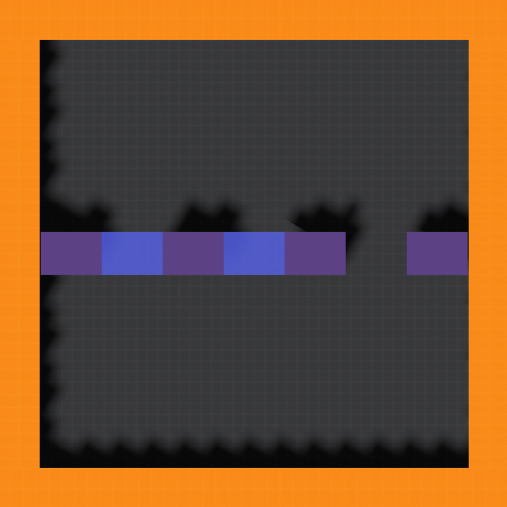}
        \subfloat{\footnotesize{S - XXO.}}
        \label{fig:envs:sxxo}
        \vspace{1em}
    \end{minipage}
    \hfill
    \begin{minipage}[h!]{0.11\linewidth}
        \centering
        \includegraphics[width=1\linewidth]{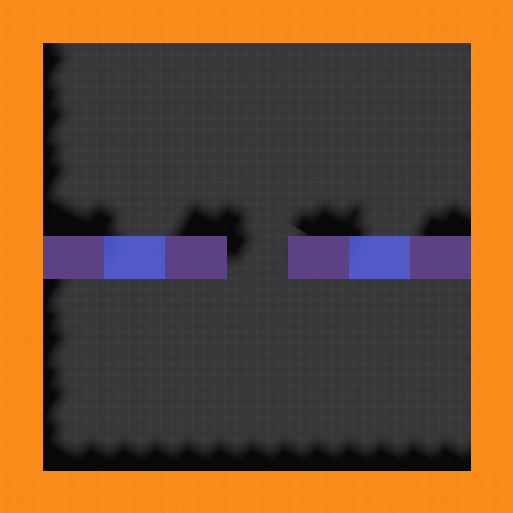}
        \subfloat{\footnotesize{S - XOX.}}
        \vspace{1em}
        \label{fig:envs:sxox}
    \end{minipage}
    \hfill
    \begin{minipage}[h!]{0.11\linewidth}
        \centering
        \includegraphics[width=1\linewidth]{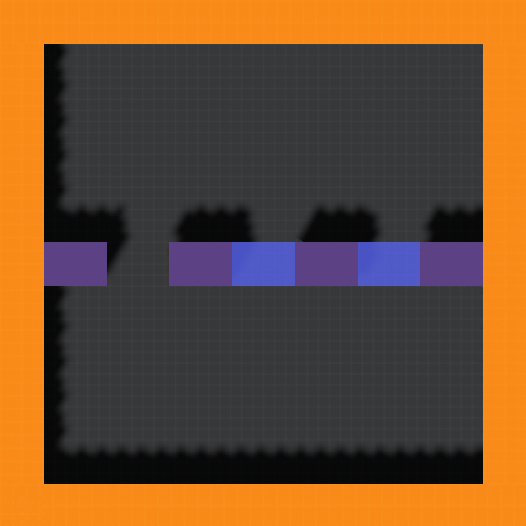}
        \subfloat{\footnotesize{S - OXX}}
        \vspace{1em}
        \label{fig:envs:soxx}
    \end{minipage}
    \begin{minipage}[h!]{0.11\linewidth}
        \centering
        \includegraphics[width=1\linewidth]{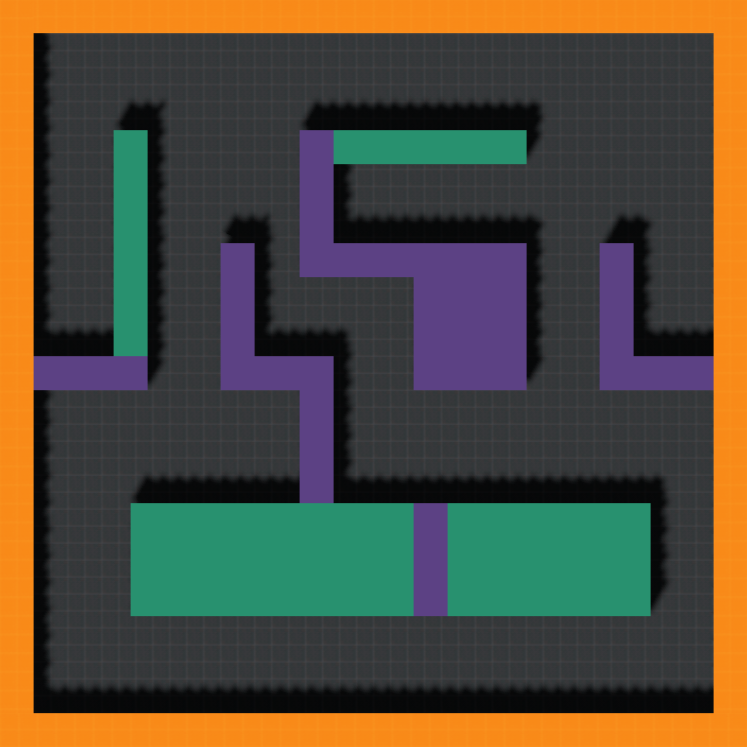}
        \subfloat{\footnotesize{A - HOOO.}}
        \label{fig:envs:ahooo}
    \end{minipage}
    \hfill
    \begin{minipage}[h!]{0.11\linewidth}
        \centering
        \includegraphics[width=1\linewidth]{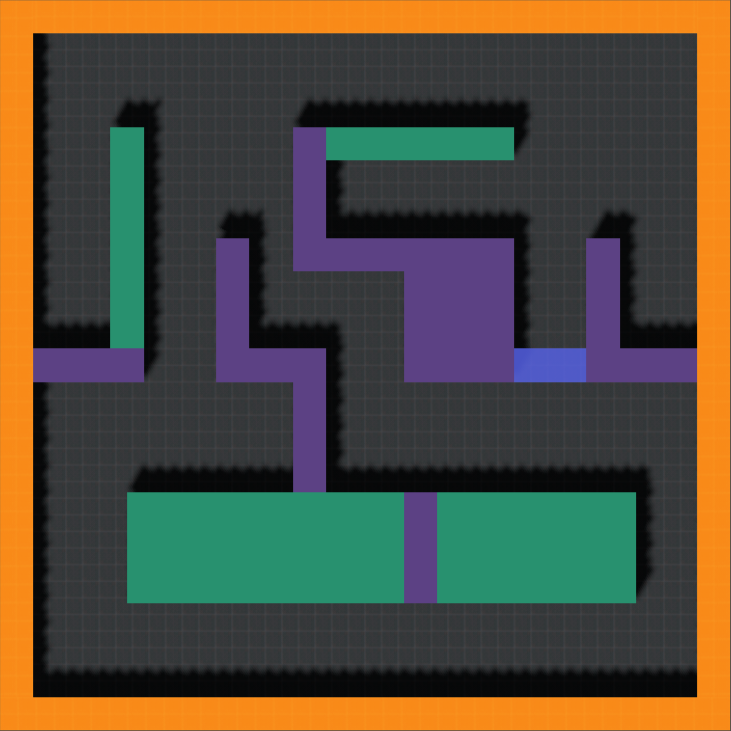}
        \subfloat{\footnotesize{A - HOOX.}}
        \label{fig:envs:ahoox}
    \end{minipage}
    \hfill
    \begin{minipage}[h!]{0.11\linewidth}
        \centering
        \includegraphics[width=1\linewidth]{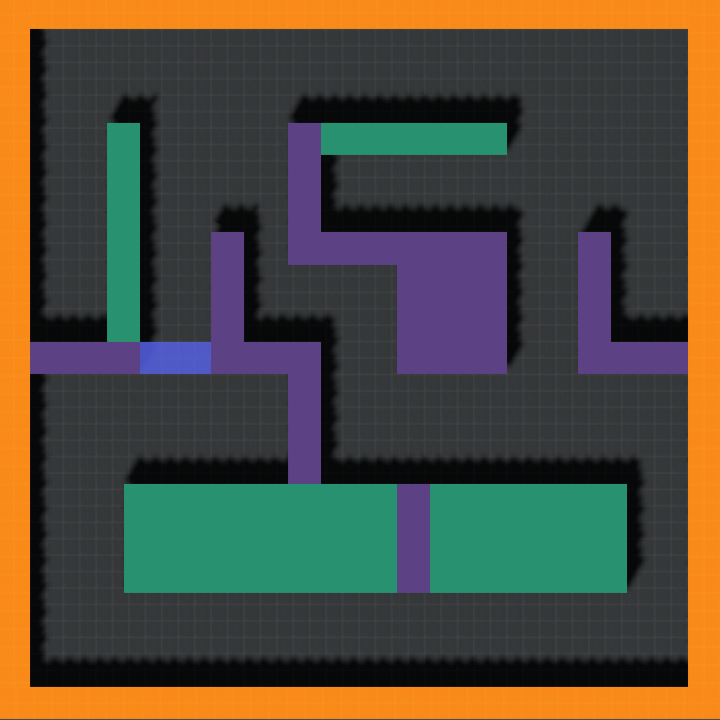}
        \subfloat{\footnotesize{A - HXOO.}}
        \label{fig:envs:ahxoo}
    \end{minipage}
    \hfill
    \begin{minipage}[h!]{0.11\linewidth}
        \centering
        \includegraphics[width=1\linewidth]{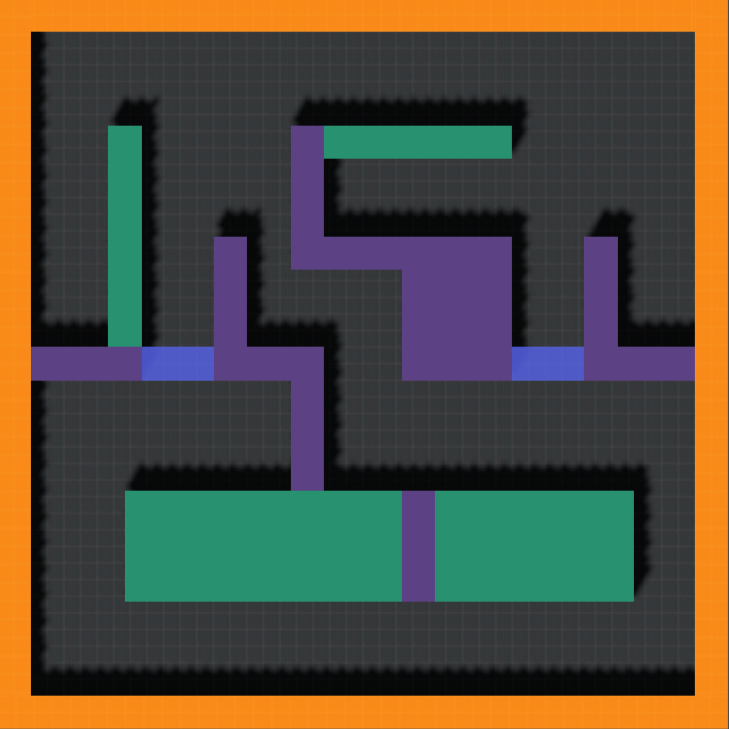}
        \subfloat{\footnotesize{A - HXOX.}}
        \label{fig:envs:ahxox}
    \end{minipage}
    \hfill
    \begin{minipage}[h!]{0.11\linewidth}
        \centering
        \includegraphics[width=1\linewidth]{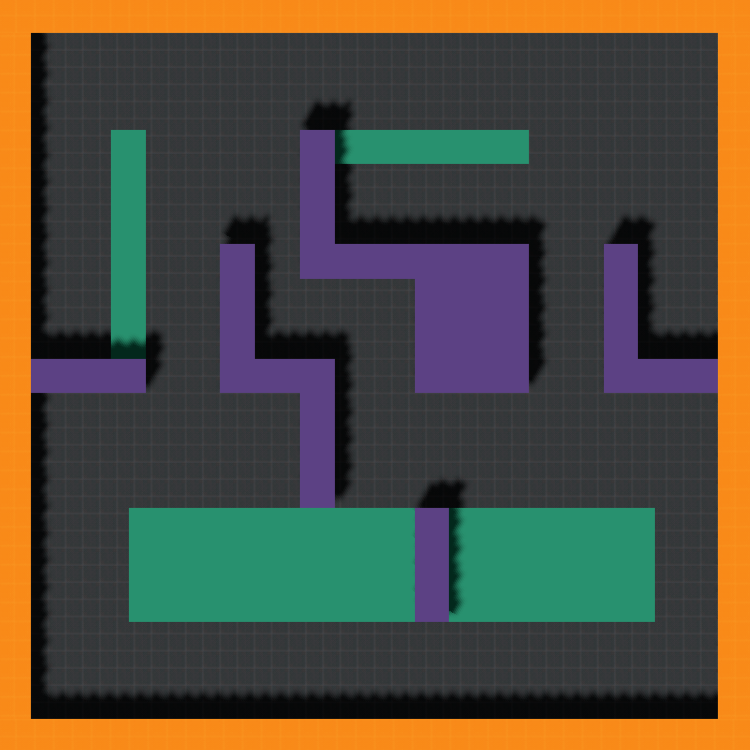}
        \subfloat{\footnotesize{A - LOOO.}}
        \label{fig:envs:alooo}
    \end{minipage}
    \hfill
    \begin{minipage}[h!]{0.11\linewidth}
        \centering
        \includegraphics[width=1\linewidth]{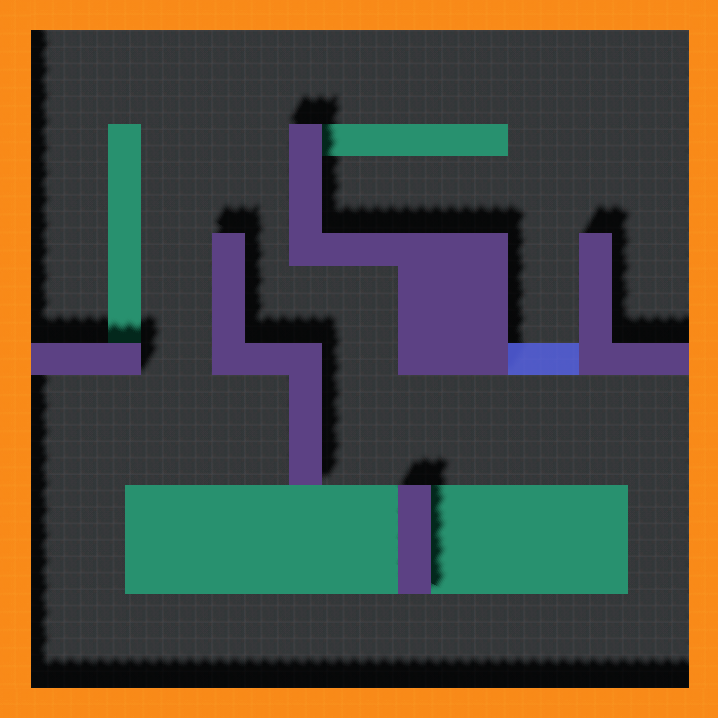}
        \subfloat{\footnotesize{A - LOOX.}}
        \label{fig:envs:aloox}
    \end{minipage}
    \hfill
    \begin{minipage}[h!]{0.11\linewidth}
        \centering
        \includegraphics[width=1\linewidth]{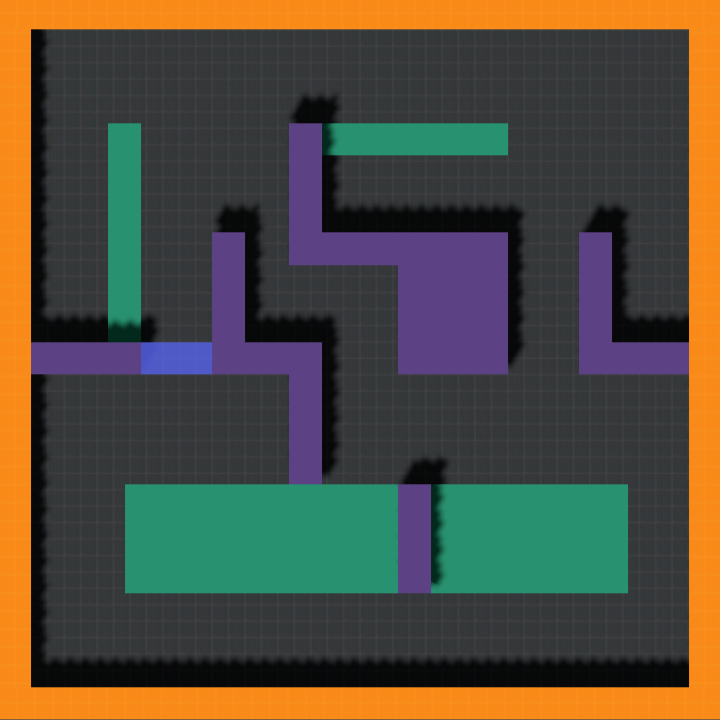}
        \subfloat{\footnotesize{A - LXOO.}}
        \label{fig:envs:alxoo}
    \end{minipage}
    \hfill
    \begin{minipage}[h!]{0.11\linewidth}
        \centering
        \includegraphics[width=1\linewidth]{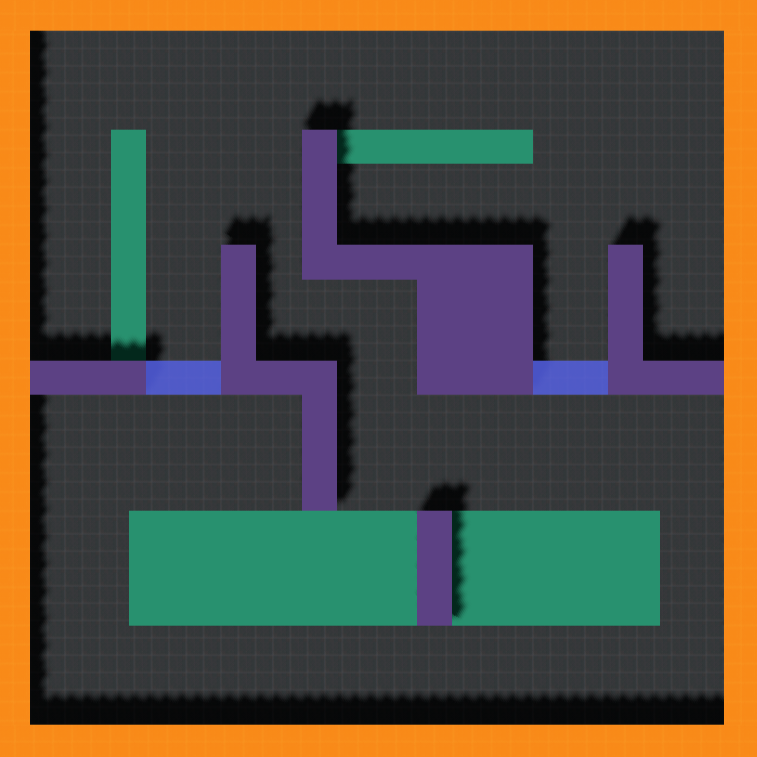}
        \subfloat{\footnotesize{A - LXOX.}}
        \label{fig:envs:}
    \end{minipage}\\
\vspace{0.25em}
\caption{\small
\textbf{SimpleTown} mazes are relatively simple, with a size of $20\times20$ meters. The starting positions are randomly sampled on one side, and the goal positions are on the other side ; \textbf{AmazeVille} mazes, of $60\times60$ meters, are more challenging. They have a finite set of start and goal positions, with two subsets of maps : some with high blocks, \textit{i.e.} not jumpable obstacles ; others with low blocks, \textit{i.e.} jumpable ones.
\textit{The naming convention of the tasks use a prefix for the maze family (S for SimpleTown, and A for AmazeVille), the subsequent characters encode key layout features where “O” and “X” are respectively open and closed doors, and “H” and “L” denote high and low blocks.}
}
\label{fig:envs:all}
\vspace{-1em}
\end{figure*}

%% file: 2-MAIN/TABLES/godot-obs-features.tex
\begin{table}[ht]

    \scriptsize
    \centering
    \vspace{0.3em}
    \begin{tblr}{
        colspec = |c|c|c||c|c|c|,
        row{1} = {Dandelion},
    }
    
        \hline
        \textbf{Feature} & \textbf{Size} & \textbf{Type} &
        \textbf{Feature} & \textbf{Size} & \textbf{Type} \\
        \hline
        \hline
        
        Position & $3$ & \texttt{float} & Orientation & $3$ & \texttt{float} \\
        Goal Position & $3$ & \texttt{float} & Velocity & $3$ & \texttt{float} \\
        RGB Image & $(3,64,64)$ & \texttt{float} & Depth Image & $(11,11)$ & \texttt{float} \\
        Floor Contact & $1$ & \texttt{bool} & Wall Contact & $1$ & \texttt{bool} \\
        Goal Contact & $1$ & \texttt{bool} & Timestep  & $1$ & \texttt{int} \\
        Up Direction & $3$ & \texttt{float} & - & - & - \\

        \hline
        
    \end{tblr}
    \caption{\small\textbf{Available observation features.} \textit{Position} data are in $m$, while the \textit{Orientation} is in $rad$, and the \textit{Velocity} in $m\cdot s^{-1}$. The \textit{RGB Images} offer a visualization of the agent's viewpoint, and the \textit{Depth Image} is generated from $11\times11$ raycasts thrown towards nearest obstacles. Although our experiments only use positions and depth images, we also provide the pixel-based observations with all datasets.}
    \label{fig:envs:godot-obs-features}
    \vspace{-2em}

\end{table}

%% file: 2-MAIN/2-baselines.tex
\section{Baseline Details}
\label{sec:baseline}

\subsection{Hierarchical Imitation Learning as Backbone Algorithm}

We consider Hierarchical Imitation Learning (HGCBC) \cite{h_bc} to address GCRL by decomposing tasks into more manageable ones. Given a MDP $\mathcal{M} = \bigl(\mathcal{S}, \mathcal{A}, \mathcal{P}_{\mathcal{S}}, {\mathcal{P}_{\mathcal{S},\mathcal{G}}}^{(0)}, \mathcal{R}, \gamma, \mathcal{G}, \phi, d\bigr)$ and a pre-collected dataset $\mathcal{D} = \{(s_t^i, a_t^i, r_t^i, s_{t+1}^i, g^i)\}$,
the end-to-end policy \(\pi_\theta(s, g)\) is split into two components :
\vspace{0.25em}
\begin{itemize}[leftmargin=*]
    \item A \textbf{High-Level Policy \(\pi^h_{\theta_h}(s, g)\)}, which selects intermediate sub-goals that serve as waypoints towards the final goal ;
    \vspace{0.25em}
    \item A \textbf{Low-Level Policy \(\pi^l_{\theta_l}(s, g)\)}, which generates actions to move the agent towards the provided sub-goal.
\end{itemize}
\vspace{0.25em}
The high-level policy is optimized by minimizing :
$$
\mathcal{L}_{\mathcal{D}}^h(\theta_h) = \mathbb{E}_{(s_t^i, s_{t+k}^i, g^i)\sim\mathcal{D}} \Bigl[-\log\bigl(\pi^h_{\theta_h}(\phi(s_{t+k}^i) \mid s_t^i, g^i)\bigr)\Bigr],
$$
while the low-level policy minimizes :
$$
\mathcal{L}_{\mathcal{D}}^l(\theta_l) = \mathbb{E}_{(s_t^i, a_t^i, s_{t+1}^i, s_{t+k}^i, g^i)\sim\mathcal{D}} \Bigl[-\log\bigl(\pi^l_{\theta_l}(a_t^i \mid s_t^i, \phi(s_{t+k}^i))\bigr)\Bigr].
$$
Hence, the hyperparameter $k\in\mathbb{N}^*$ defines the temporal gap.
\vspace{-0.5em}
\begin{figure}[h]
    \centering
    \begin{minipage}[h!]{0.49\linewidth}
        \centering
        \includegraphics[width=1\linewidth]{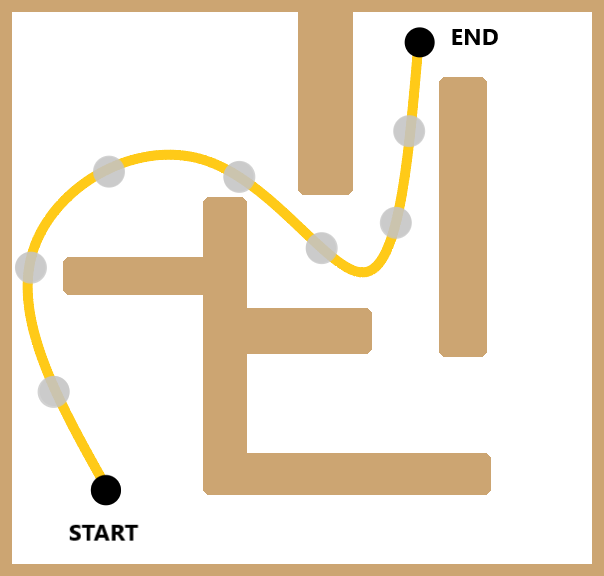}
        \subfloat{\footnotesize(A) Original Trajectory.}
        \label{fig:her:left}
    \end{minipage}
    \hfill
    \begin{minipage}[h!]{0.49\linewidth}
        \centering
        \includegraphics[width=1\linewidth]{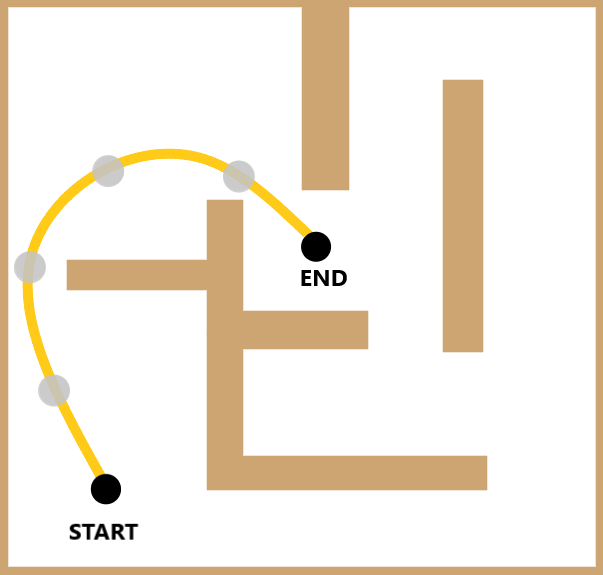}
        \subfloat{\footnotesize(B) Relabeled Trajectory.}
        \label{fig:her:right}
    \end{minipage}
\caption{\small
\textbf{Hindsight Experience Replay (HER) Illustration.}}
\label{fig:her}
\vspace{-0.75em}
\end{figure}
\textit{Figure ~\ref{fig:her}} illustrates \textit{Hindsight Experience Replay (HER)} as a data augmentation strategy, leveraging the hierarchical structure. 
By relabeling transitions with alternative sub-goals, HER can enrich the dataset, especially in low or sparse data regimes.

\subsection{Continual Reinforcement Learning Baselines}

We evaluate a range of baseline methods designed to learn from a task stream. 
These methods span several categories :\\

{
\setlength{\parindent}{0pt}
\vspace{-0.5em}
\underline{\textbf{Naive Methods :}}\\

\vspace{-0.5em}
\textbf{Naive Learning Strategy or \textit{From Scratch} (SC1 \& SCN) :}
in SC1, a single policy is learned from the latest dataset and then applied unchanged to all tasks. 
In SCN, a new policy is trained for each task, improving performance at the cost of a memory load.
\textbf{Freeze Strategy (FRZ) :} here, a single policy is trained on the first task and applied without modification to all subsequent tasks.
\textbf{Finetuning Strategy (FT1 \& FTN) :} the Finetuning Strategy involves adapting a policy learned from the initial task to each subsequent task, either by continuously updating a single policy (FT1) or by copying and then updating the policy for each new task (FTN), allowing for a better adaptation.

\underline{\textbf{Replay-Based Method :}}\\

\vspace{-0.5em}
\textbf{Experience Replay Strategy (RPL) :} this method aggregates all task datasets, up to the current one, into a single replay buffer and trains a single policy on the combined data, ensuring continuous exposure to earlier experiences. 
Such a process may allow to avoid catastrophic forgetting, and eventually build common skills and knowledge accross tasks.\\

\underline{\textbf{Weight Regularization Methods :}}\\

\vspace{-0.5em}
\textbf{Elastic Weight Consolidation (EWC) \cite{ewc} :} this strategy mitigates catastrophic forgetting by selectively slowing down learning on certain weights based on their importance to previously learned tasks. 
This importance is measured by the Fisher Information Matrix, which quantifies the sensitivity of the output function to
changes in the parameters.
EWC introduces a quadratic penalty, constraining the parameters close to their values from previous tasks, where the strength of the penalty is proportional to each parameter’s importance.
\textbf{L2-Regularization Finetuning (L2) \cite{l2reg} :} This strategy also mitigates catastrophic forgetting by adding an L2 penalty to the loss, discouraging large weight changes during training.
This helps preserve knowledge from previous tasks by promoting stability in the learned representations.
As with EWC, L2-regularization struggles in CRL for navigation tasks, especially when actions or dynamics change drastically. 
The method limits the network’s flexibility by forcing small weight updates, making it difficult to adapt to tasks that require distinct actions for similar states, which is critical in evolving environments.\\

\underline{\textbf{Architectural Methods :}}\\

\vspace{-0.5em}
\textbf{Progressive Neural Networks (PNN) \cite{pnn} :} This framework introduce a new column layer for each task, freezing previous weights to preserve knowledge. Lateral connections allow feature transfer, leveraging prior experience while avoiding interference.
By integrating previously learned features, PNN builds richer compositional representations, ensuring that prior knowledge is preserved and used throughout the feature hierarchy.
Nevertheless the model grows with each task, limiting scalability for long streams tasks or limited memory contexts.
\textbf{Hierarchical Subspace of Policies (HiSPO) \cite{hispo}:} this strategy partitions the model into distinct subspaces of neural networks \cite{sub_nnsub} for high-level and low-level control. Initially, anchor parameters for both subspaces are trained on the first task to form a foundational knowledge reservoir. For each subsequent task, new anchors are introduced and optimized using the task’s dataset. The algorithm then samples anchor weight configurations to explore the previous subspace, and it selects the configuration that minimizes the loss. Finally, by comparing the loss of the extended subspace with that of the previous configuration, under a threshold $\epsilon$, HiSPO either prunes the new anchor or retains it. This process integrates new information while preserving and reusing prior knowledge, ensuring scalability and adaptability across tasks.
}

\subsection{Implementation Details}

{
\setlength{\parindent}{0pt}

We base our configuration on the approach described in \cite{passive_data}. In all our experiments, we use Residual MLPs \cite{res} with Layer Normalization \cite{layernorm} for the hidden layers. All networks use\linebreak
$3\times256$ hidden units, GELU activations, and are initialized with a variance scaling strategy \cite{var_init} (with a scale of $0.1$).\\

\vspace{-0.6em}
We set the batch size to $64$, the learning rate to $3\times10^{-4}$, the number of gradient steps to $1\times10^5$, and the sub-goal distance to $k=10$ for AmazeVille and $k=5$ for SimpleTown. 
HER sampling temperatures are respectively set to $100.0$ and $15.0$.\\

\vspace{-0.6em}
For both EWC and L2, we considered five regularization strengths 
$\lambda \in \{1\times10^{-2}, 1\times10^{-1}, 1, 1\times10^{1}, 1\times10^{2}\}$ and selected the best-performing model for each stream. Similarly, for HiSPO, we experimented with acceptance thresholds 
$\epsilon_h,\epsilon_l \in \{5\times10^{-2}, 1\times10^{-1}\}$ and weights for the similarity loss as $\lambda_h,\lambda_l \in \{5\times10^{-2}, 1\times10^{-1}\}$, for functional diversity.\\

\vspace{-0.6em}
We used a compute cluster featuring Intel(R) Xeon(R) CPU E5-1650 and Intel Cascade Lake 6248 processors. For most models, 4 cores were sufficient, but due to PNN’s growing memory requirements, we allocated 6 cores for its experiments.
}

%% file: 2-MAIN/3-benchmark.tex
\section{Benchmark Study}
\label{sec:benchmark}

\subsection{Offline Reinforcement Learning}

{
\setlength{\parindent}{0pt}

We first compare the hierarchical backbone (HGCBC) with a Goal-Conditioned Behavioral Cloning (GCBC) \cite{gcbc} baseline, which is the non-hierarchical supervised baseline approach.

}

\input{2-MAIN/TABLES/1_gcbc_vs_hgcbc}

\subsection{Continual Offline Reinforcement Learning}

{
\setlength{\parindent}{0pt}

\subsubsection{Performance }

\textit{Table \ref{fig:bench:perf}} presents the performance metric of each CRL method on the task streams. To a certain extent, PER simultaneously tells about generalization and forgetting.

\input{2-MAIN/TABLES/2a_performance}

\subsubsection{Backward \& Forward Transfer }

\textit{Figure \ref{fig:bench:bwt_fwt}} presents the BWT (\ref{fig:bench:bwt_fwt}.A) and FWT (\ref{fig:bench:bwt_fwt}.B) metrics, thus focusing on the generalization and forgetting phenomena across the streams.

\begin{figure}[h]
    \vspace{-0.125em}
    \centering
    \begin{minipage}[h!]{\linewidth}
        \centering
        \includegraphics[width=\linewidth]{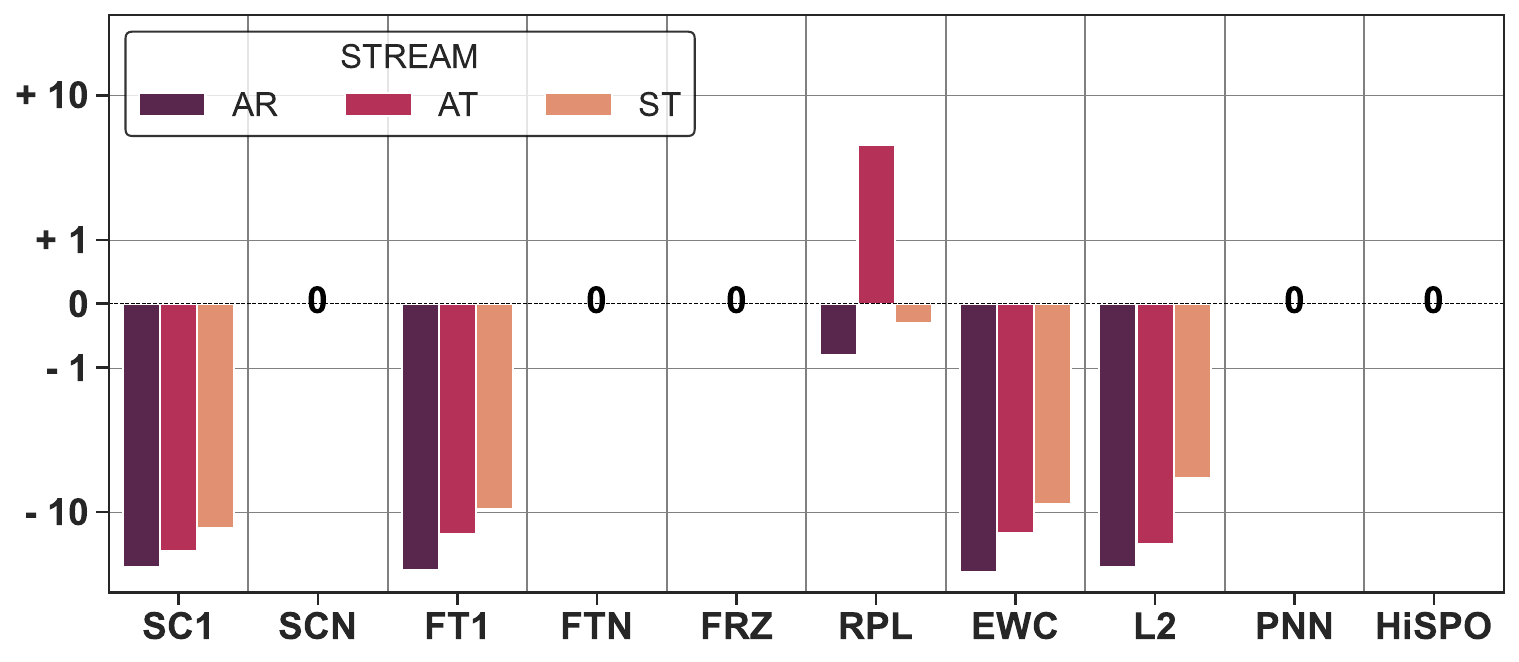}
        \subfloat{\footnotesize(A) Backward Transfer Metric.}
        \label{fig:bench:bwt}
        \vspace{0.75em}
    \end{minipage}
    \hfill
    \begin{minipage}[h!]{\linewidth}
        \centering
        \includegraphics[width=\linewidth]{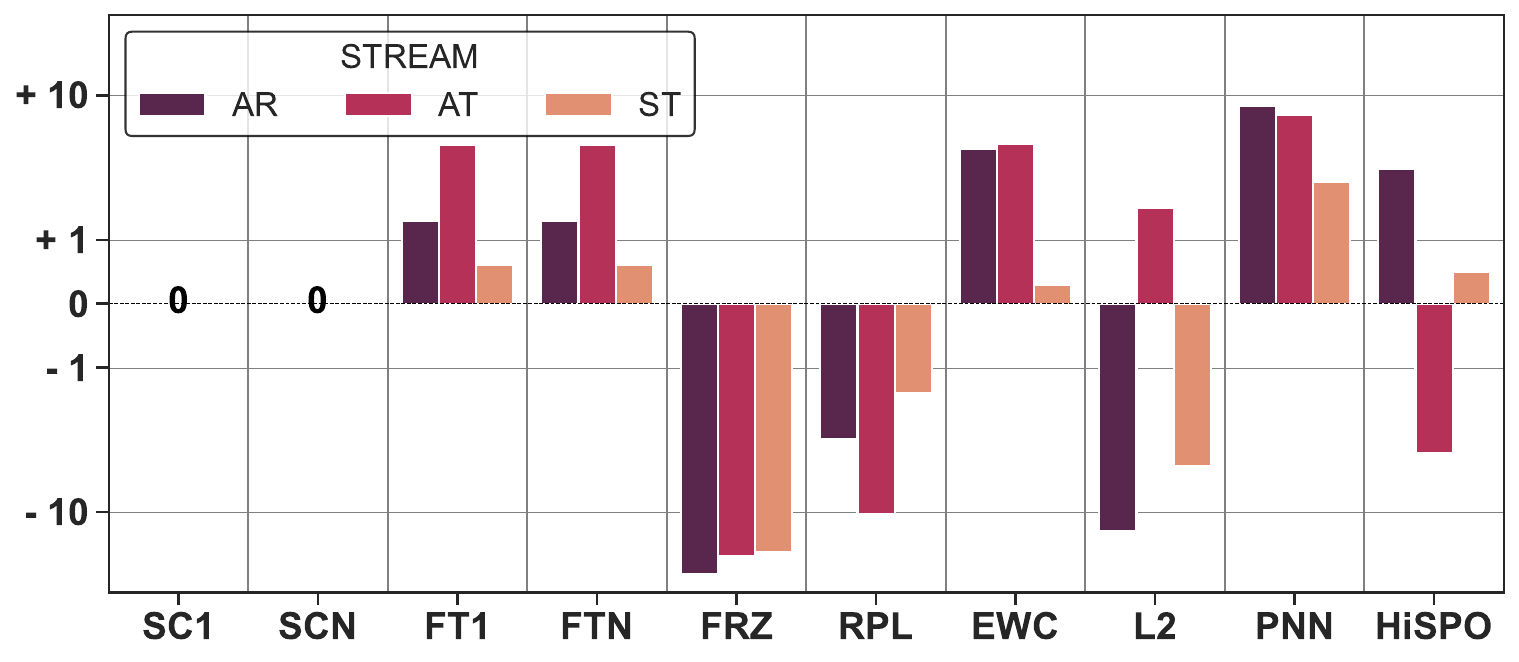}
        \subfloat{\footnotesize(B) Forward Transfer Metric.}
        \label{fig:bench:fwt}
        \vspace{0.25em}
    \end{minipage}
\caption{\small
\textbf{Backward and Forward Transfer Metrics.}
Architectural or separate‐policy methods (\textit{PNN}, \textit{HiSPO}, \textit{SCN}, \textit{FTN}) typically balance old‐task retention and new‐task adaptation, whereas single‐policy or regularized methods (\textit{SC1}, \textit{FRZ}, \textit{EWC}, \textit{L2}) risk higher forgetting or reduced forward transfer in complex navigation scenarios.
}
\label{fig:bench:bwt_fwt}
\end{figure}

\subsubsection{Relative Model Size }

On one hand, Figure~\ref{fig:bench:mem_inference} shows how much each model grows relative to a reference network (\textit{SC1}) at inference time. This metric is crucial for production settings, with storage or runtime overheads to minimize.
On the other hand, Figure~\ref{fig:bench:mem_training} shows the relative size during training. This metric is relevant to track the scalability of a learning method.

\begin{figure}[h]
    \centering
    \begin{minipage}[h!]{\linewidth}
        \centering
        \includegraphics[width=1\linewidth]{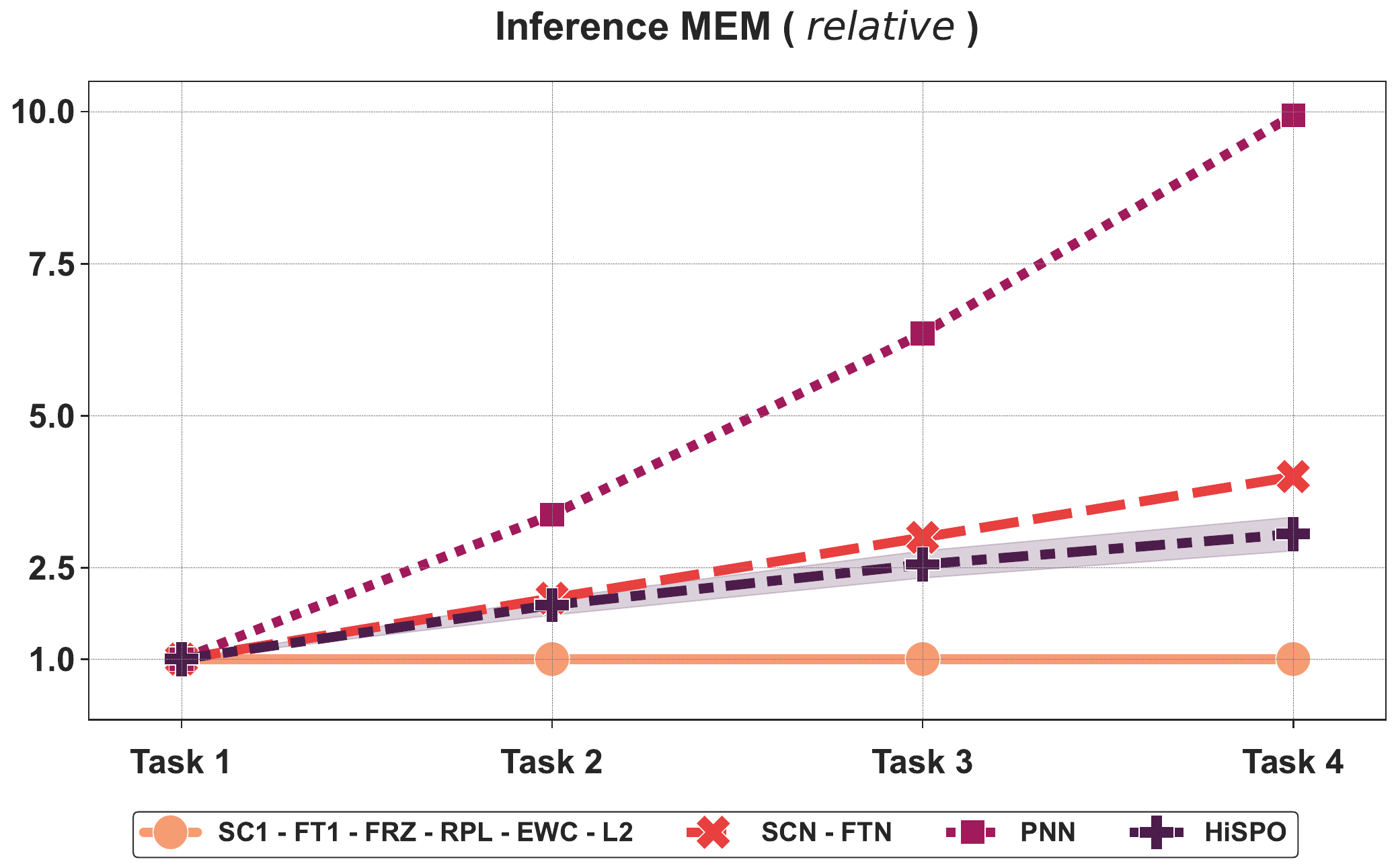}
    \end{minipage}
\vspace{1em}
\caption{\small
\textbf{Average Relative MEM Metric \textit{at Inference}.}
\textbf{PNN} exhibits the fastest growth, as each new task introduces an additional column and layer interconnections to all previously learned features maps.\linebreak
\textbf{SCN} and \textbf{FTN} grow linearly, allocating a new policy for each new task. In contrast, \textbf{SC1}, \textbf{FRZ}, \textbf{EWC}, \textbf{L2}, and \textbf{RPL} keep a single model, so their size remains constant for any stream of tasks. Finally, \textbf{HiSPO} increases subspace parameters only when necessary, leading to sublinear or moderate growth relative to purely additive approaches.
}
\label{fig:bench:mem_inference}
\end{figure}

\begin{figure}[h]
    \centering
    \begin{minipage}[h!]{\linewidth}
        \centering
        \includegraphics[width=1\linewidth]{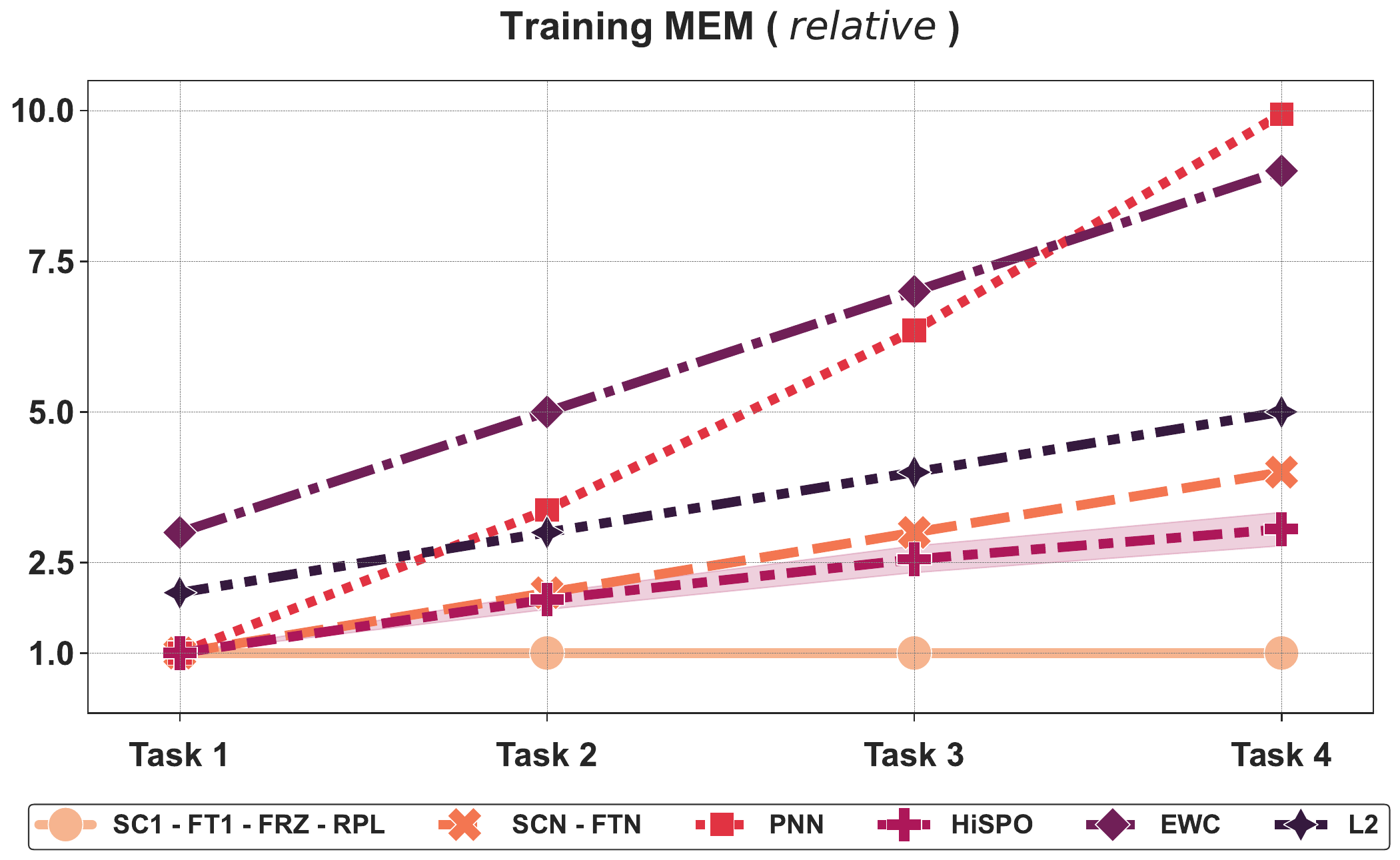}
    \end{minipage}
\vspace{1em}
\caption{\small
\textbf{Average Relative MEM Metric \textit{at Training}.}
Most methods preserve a single working model, so their training overhead remains similar to the inference one. However, \textit{EWC} and \textit{L2} store additional parameter snapshots (old weights for both, and Fisher Information for EWC) to constrain updates and reduce forgetting. This extra storage can raise memory demands when many tasks are encountered.
}
\label{fig:bench:mem_training}
\end{figure}

These results highlight a core trade‐off : methods like \textit{PNN} or \textit{SCN \& FTN} can improve continual learning but incur greater memory overhead, while all the single‐model approaches (such as \textit{SC1}, \textit{FRZ}, \textit{RPL}) remain lightweight but risk higher forgetting. Balancing memory cost and adaptation is thus key to deploying continual offline reinforcement learning in practical pipelines.

\subsubsection{Training \& Inference Costs }

Continual Reinforcement Learning methods differ not only in memory overhead but also in computational demands during both training and inference. Figure~\ref{fig:bench:time_training} reports TRN, the average models training cost metric (in minutes), and Figure~\ref{fig:bench:time_inference} shows INF, the average models inference cost metric (in ms). 

\begin{figure}[h]
    \centering
    \begin{minipage}[h!]{\linewidth}
        \centering
        \includegraphics[width=1\linewidth]{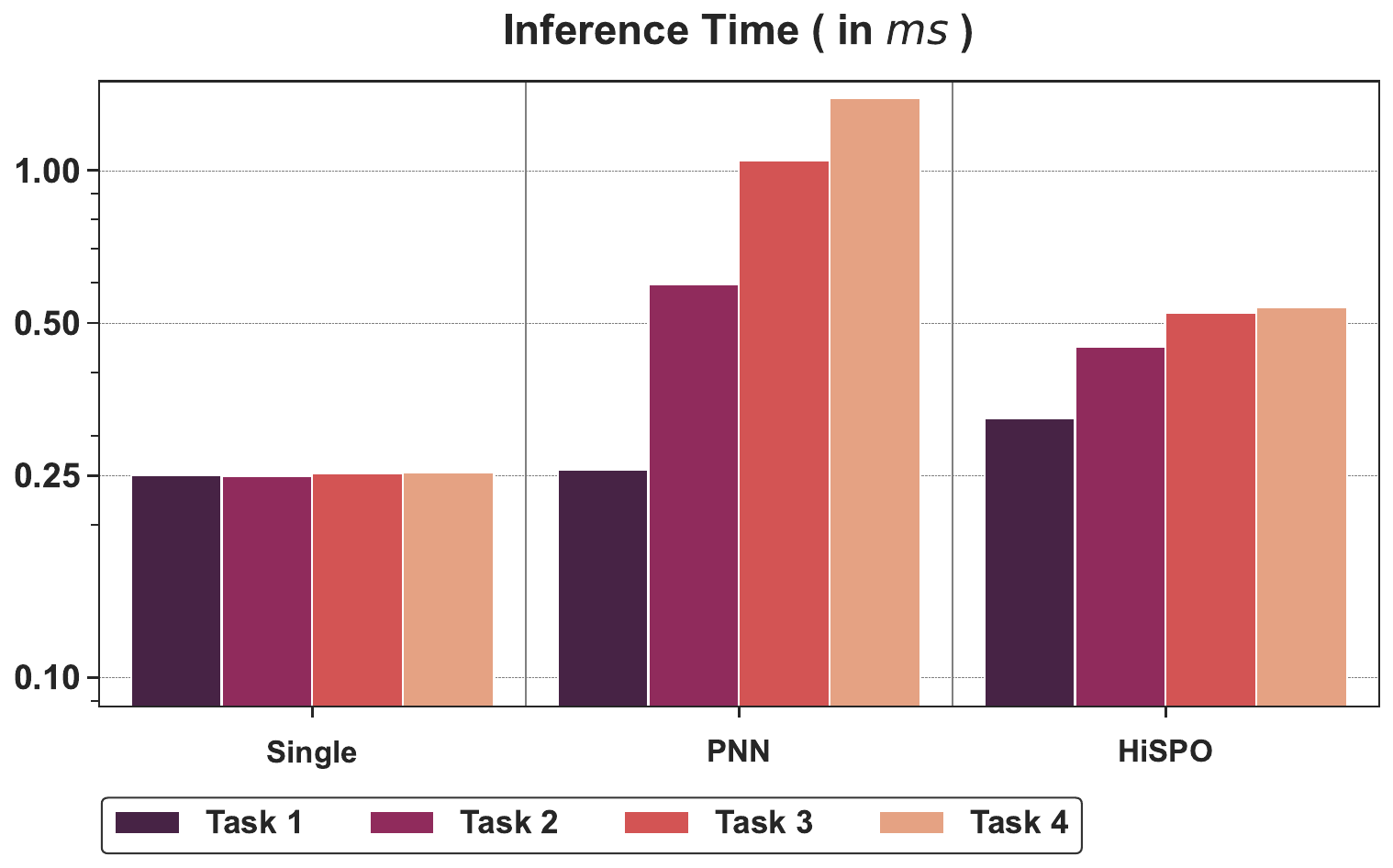}
    \end{minipage}
\vspace{1em}
\caption{\small
\textbf{Average INF (model inference cost duration \textit{in ms}).}
We measure inference time averaging on 100{,}000 forward passes on CPU with a batch size of 64, using the AR1 stream (seed 100).\linebreak
At inference, single‐model approaches (\textit{SC1-N}, \textit{FT1-N}, \textit{FRZ}, \textit{RPL}, \textit{EWC}, and \textit{L2}) incur minimal overhead, requiring a single forward pass per step. In contrast architectural strategies (\textit{PNN} and \textit{HiSPO}) may demand additional computations (such as selecting the right sub‐policy or combining anchors), leading to higher inference times.
}
\label{fig:bench:time_inference}
\end{figure}

\begin{figure}[h]
    \centering
    \begin{minipage}[h!]{\linewidth}
        \centering
        \includegraphics[width=1\linewidth]{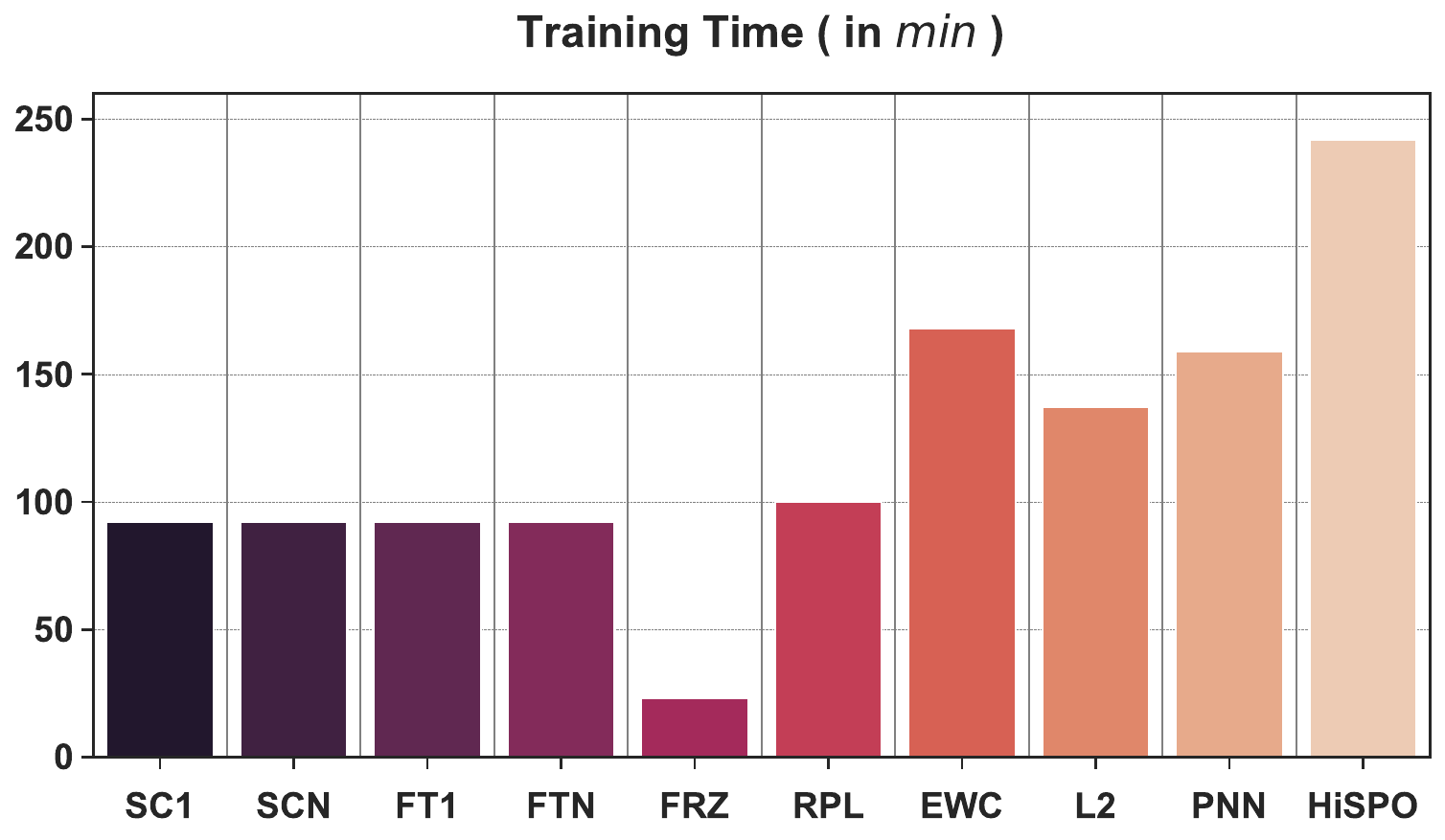}
    \end{minipage}
\vspace{0.5em}
\caption{\small
\textbf{Average TRN (training duration \textit{in min}).} Single-model approaches train quickly by updating a single parameter set, while RPL is slightly slower due to repeated dataset sampling. EWC and L2 add overhead by storing parameters snapshots. Architectural methods also take longer to train, as PNN shows non-linear growth by adding new columns and lateral connections per task, and HiSPO introduces new anchors and also computes cosine‐similarity regularization weights.
}
\label{fig:bench:time_training}
\end{figure}

Overall, these findings show that single‐model methods maintain lighter overhead during both training and inference but can suffer from higher forgetting. Conversely, multi‐policy or subspace‐based strategies better preserve knowledge at the cost of increased computational demands, a crucial consideration in large‐scale or time‐sensitive applications.

\begin{figure*}[t!]
    \centering
    \begin{minipage}[h!]{00.19\linewidth}
        \centering
        \includegraphics[width=1\linewidth]{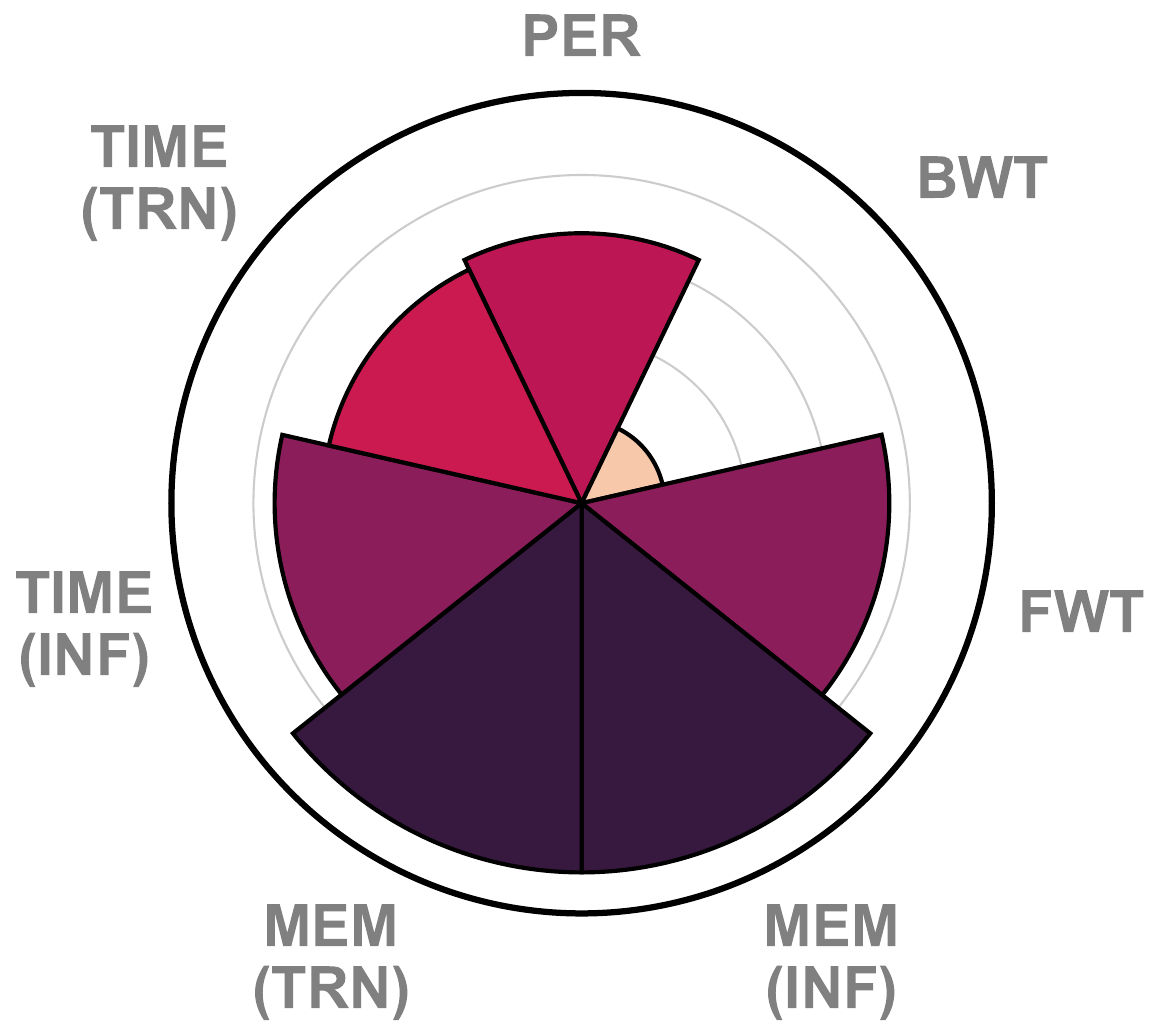}
        \subfloat{\textbf{SC1}}
        \vspace{0.75em}
        \label{fig:bench:sc1}
    \end{minipage}
    \hfill
    \begin{minipage}[h!]{00.19\linewidth}
        \centering
        \includegraphics[width=1\linewidth]{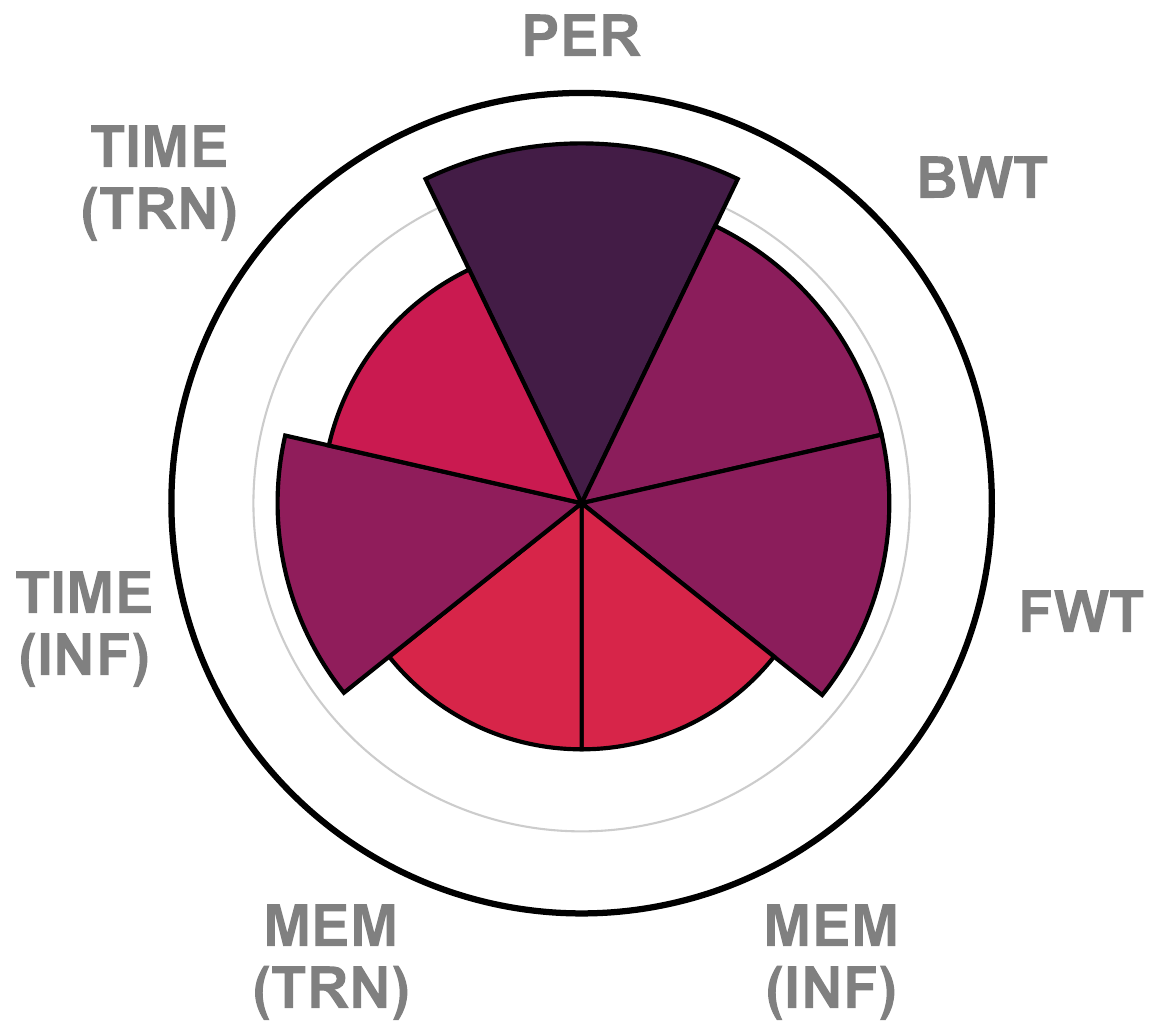}
        \subfloat{\textbf{SCN}}
        \vspace{0.75em}
        \label{fig:bench:scn}
    \end{minipage}
    \hfill
    \begin{minipage}[h!]{00.19\linewidth}
        \centering
        \includegraphics[width=1\linewidth]{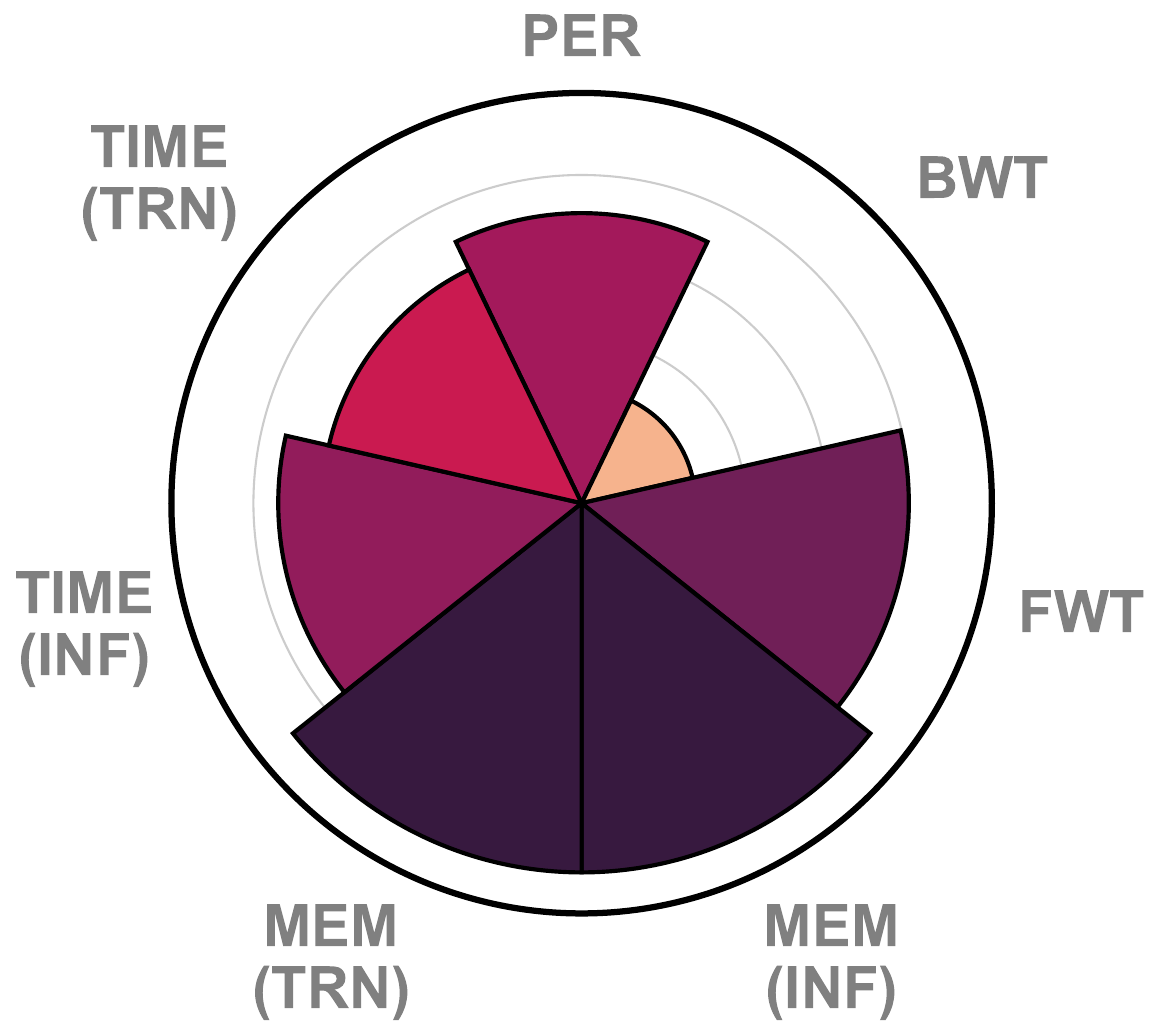}
        \subfloat{\textbf{FT1}}
        \vspace{0.75em}
        \label{fig:bench:ft1}
    \end{minipage}
    \hfill
    \begin{minipage}[h!]{00.19\linewidth}
        \centering
        \includegraphics[width=1\linewidth]{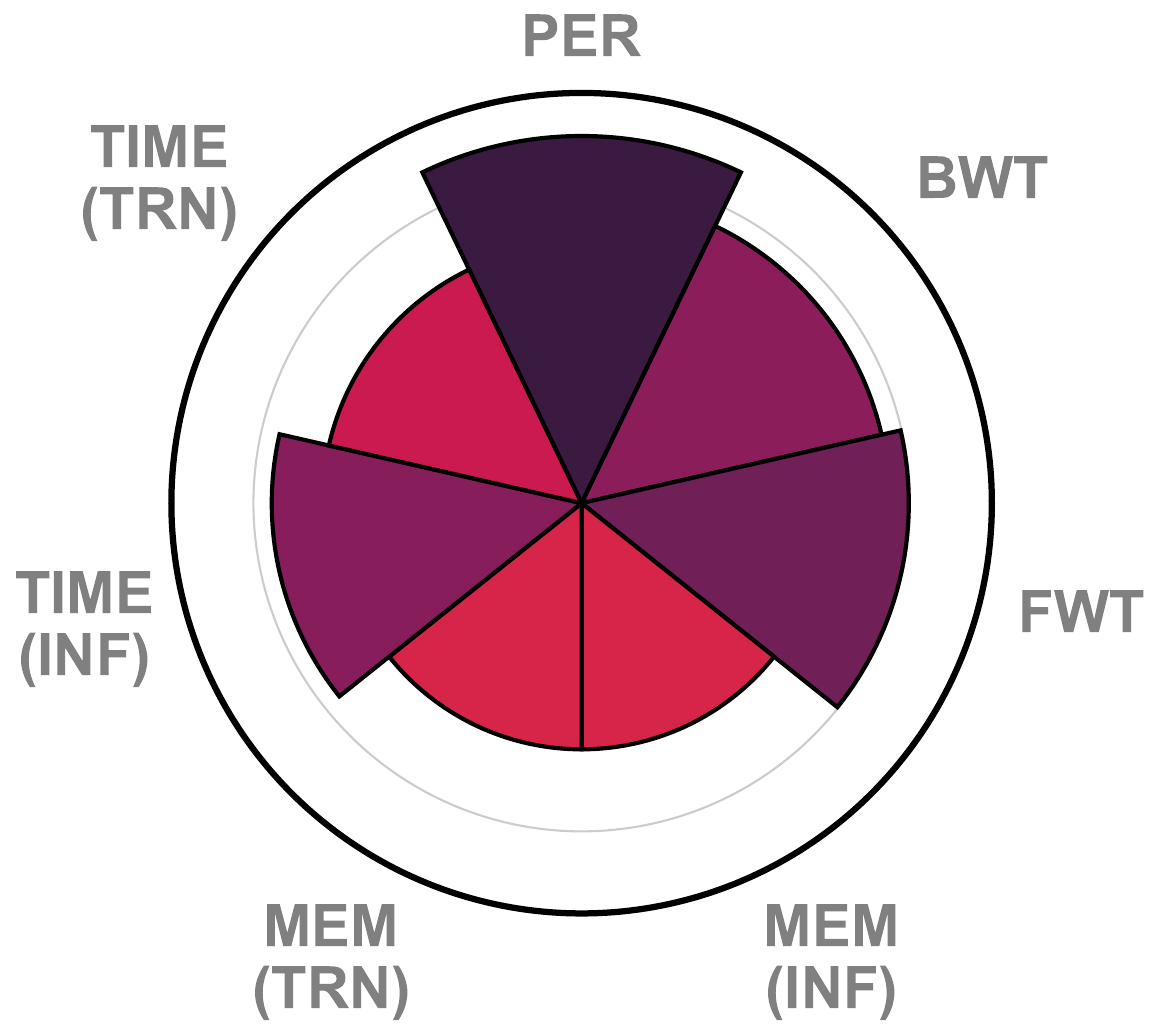}
        \subfloat{\textbf{FTN}}
        \vspace{0.75em}
        \label{fig:bench:ftn}
    \end{minipage}
    \hfill
    \begin{minipage}[h!]{00.19\linewidth}
        \centering
        \includegraphics[width=1\linewidth]{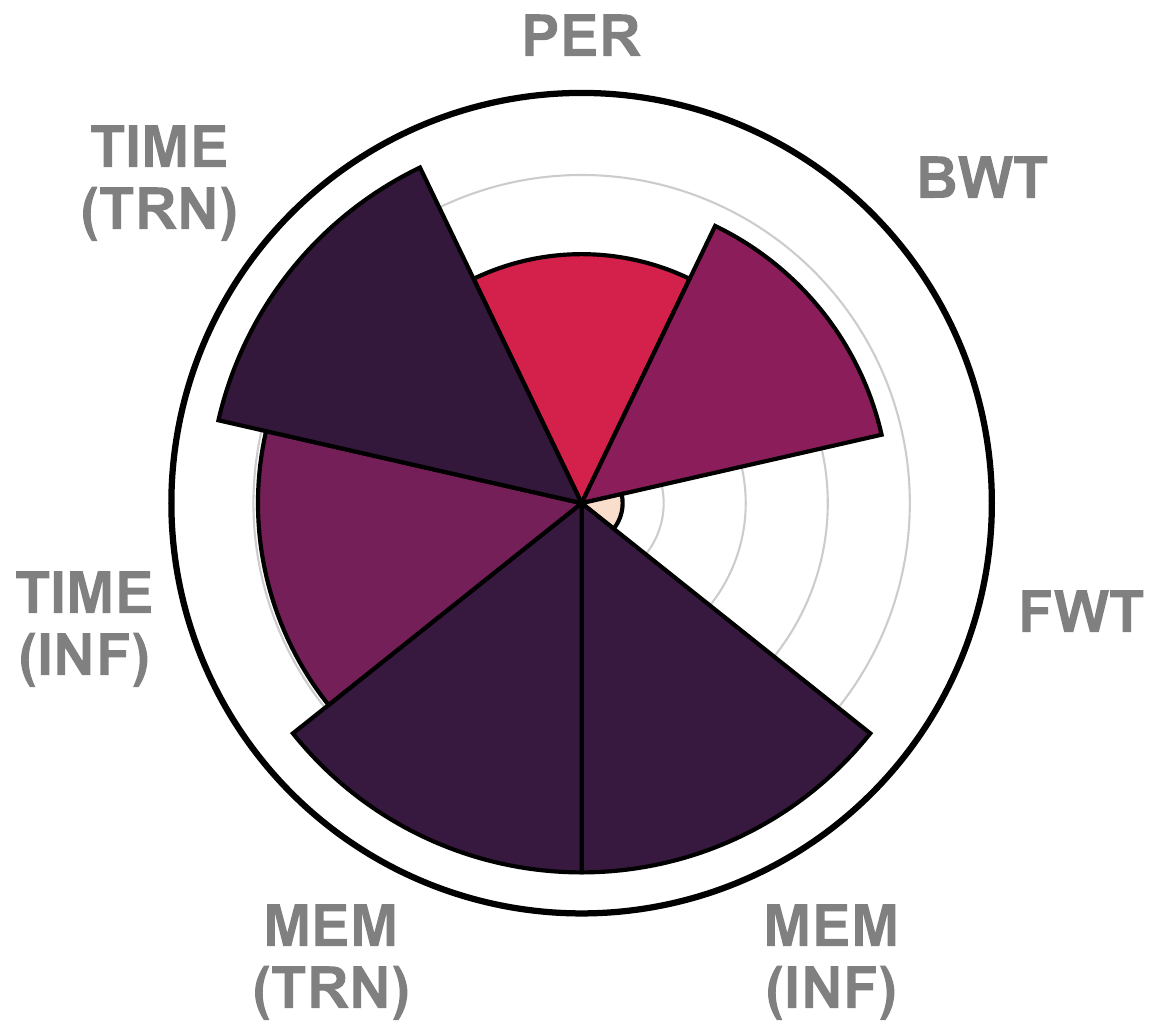}
        \subfloat{\textbf{FRZ}}
        \vspace{0.75em}
        \label{fig:bench:frz}
    \end{minipage}
    \hfill
    \begin{minipage}[h!]{00.19\linewidth}
        \centering
        \includegraphics[width=1\linewidth]{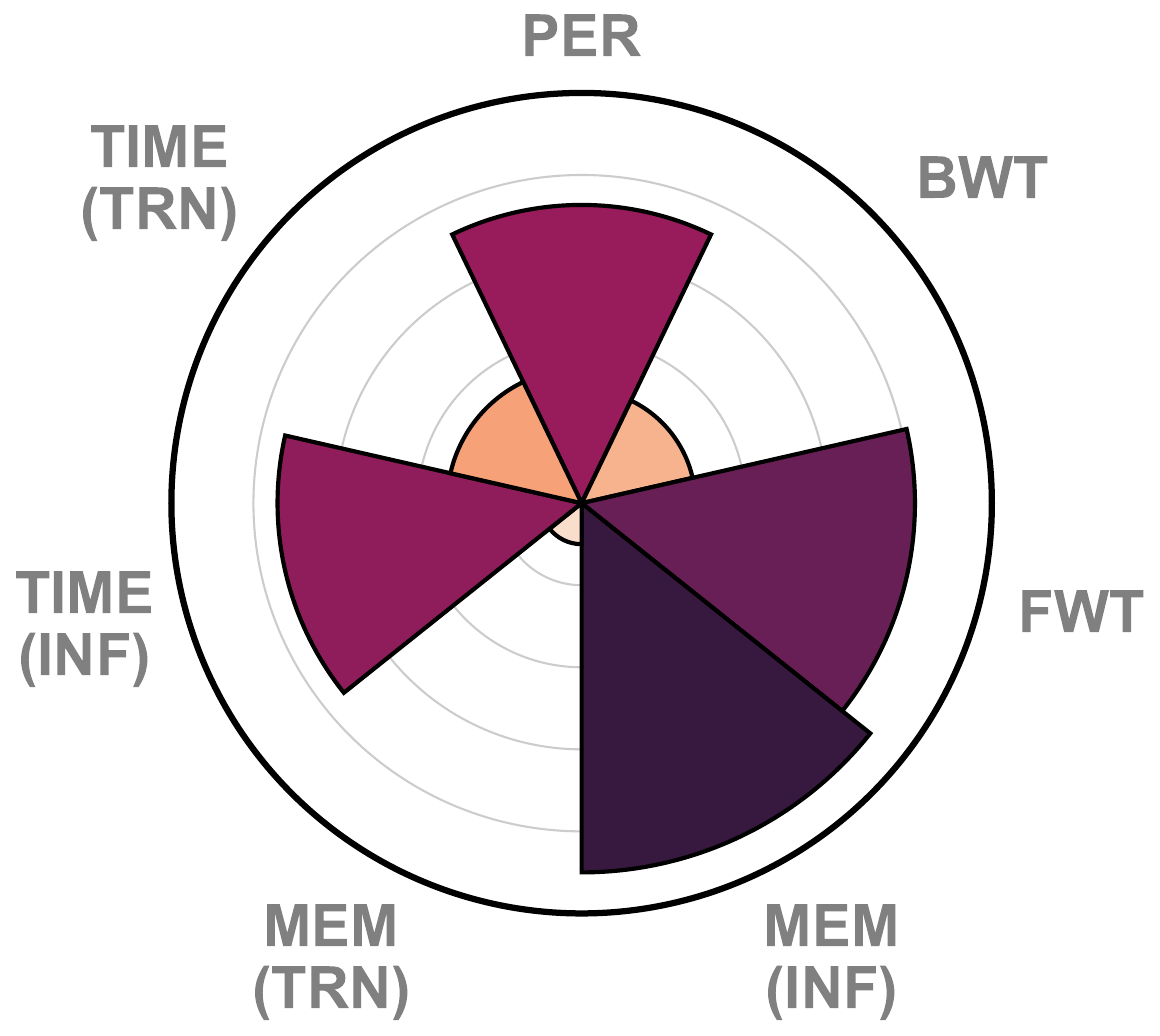}
        \subfloat{\textbf{EWC}}
        \vspace{0.25em}
        \label{fig:bench:ewc}
    \end{minipage}
    \hfill
    \begin{minipage}[h!]{00.19\linewidth}
        \centering
        \includegraphics[width=1\linewidth]{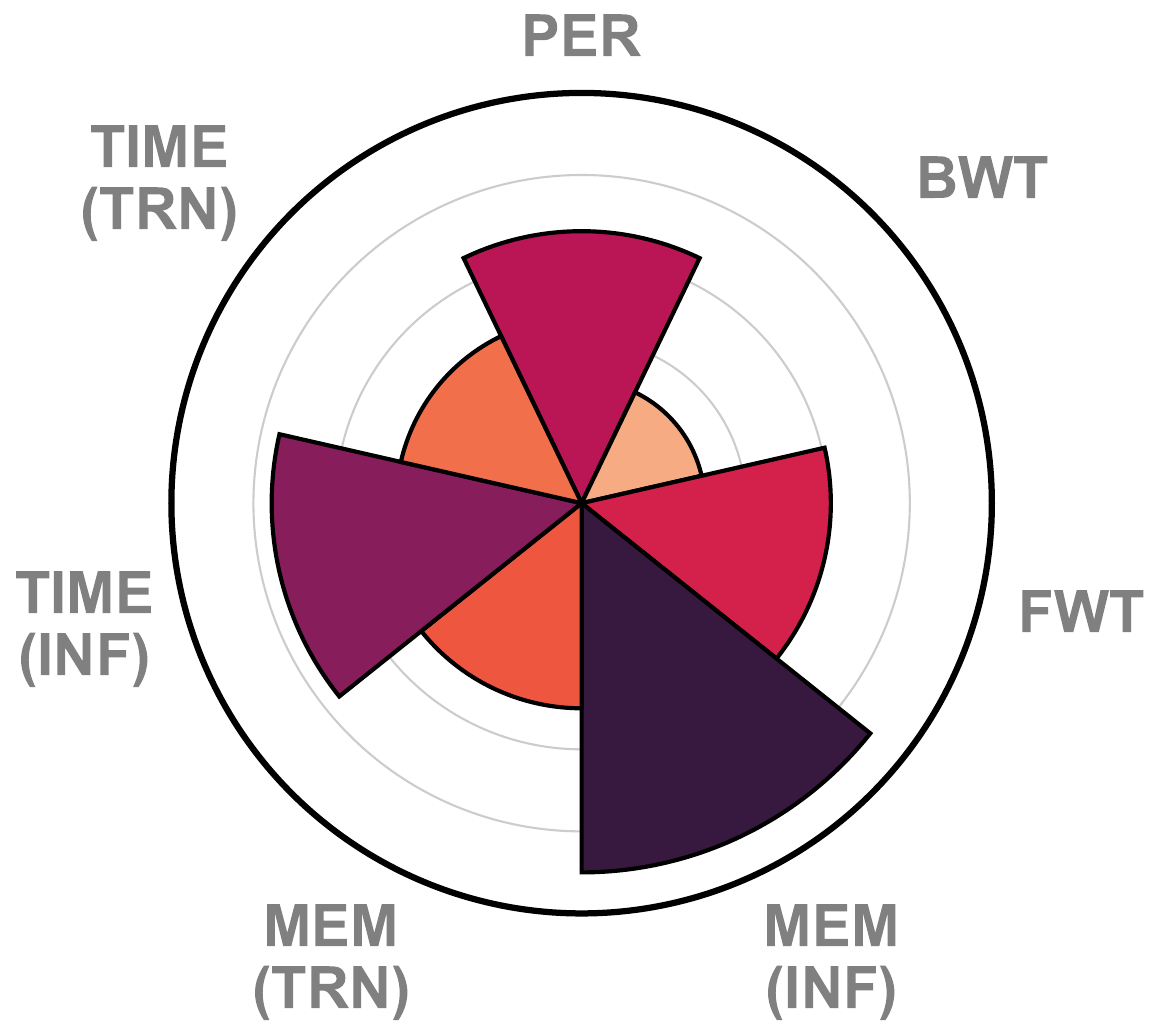}
        \subfloat{\textbf{L2}}
        \vspace{0.25em}
        \label{fig:bench:l2}
    \end{minipage}
    \hfill
    \begin{minipage}[h!]{00.19\linewidth}
        \centering
        \includegraphics[width=1\linewidth]{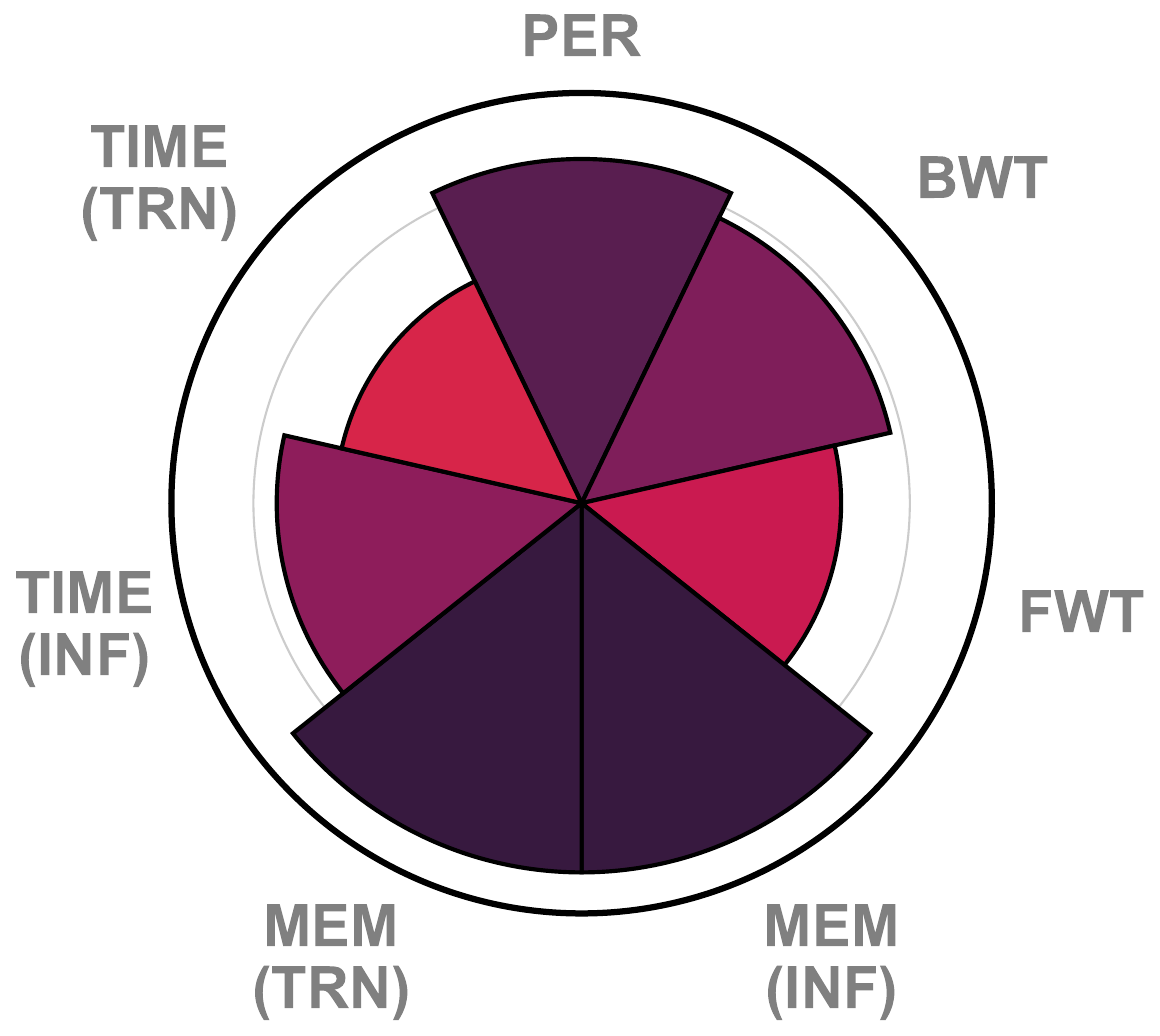}
        \subfloat{\textbf{RPL}}
        \vspace{0.25em}
        \label{fig:bench:rpl}
    \end{minipage}
    \hfill
    \begin{minipage}[h!]{00.19\linewidth}
        \centering
        \includegraphics[width=1\linewidth]{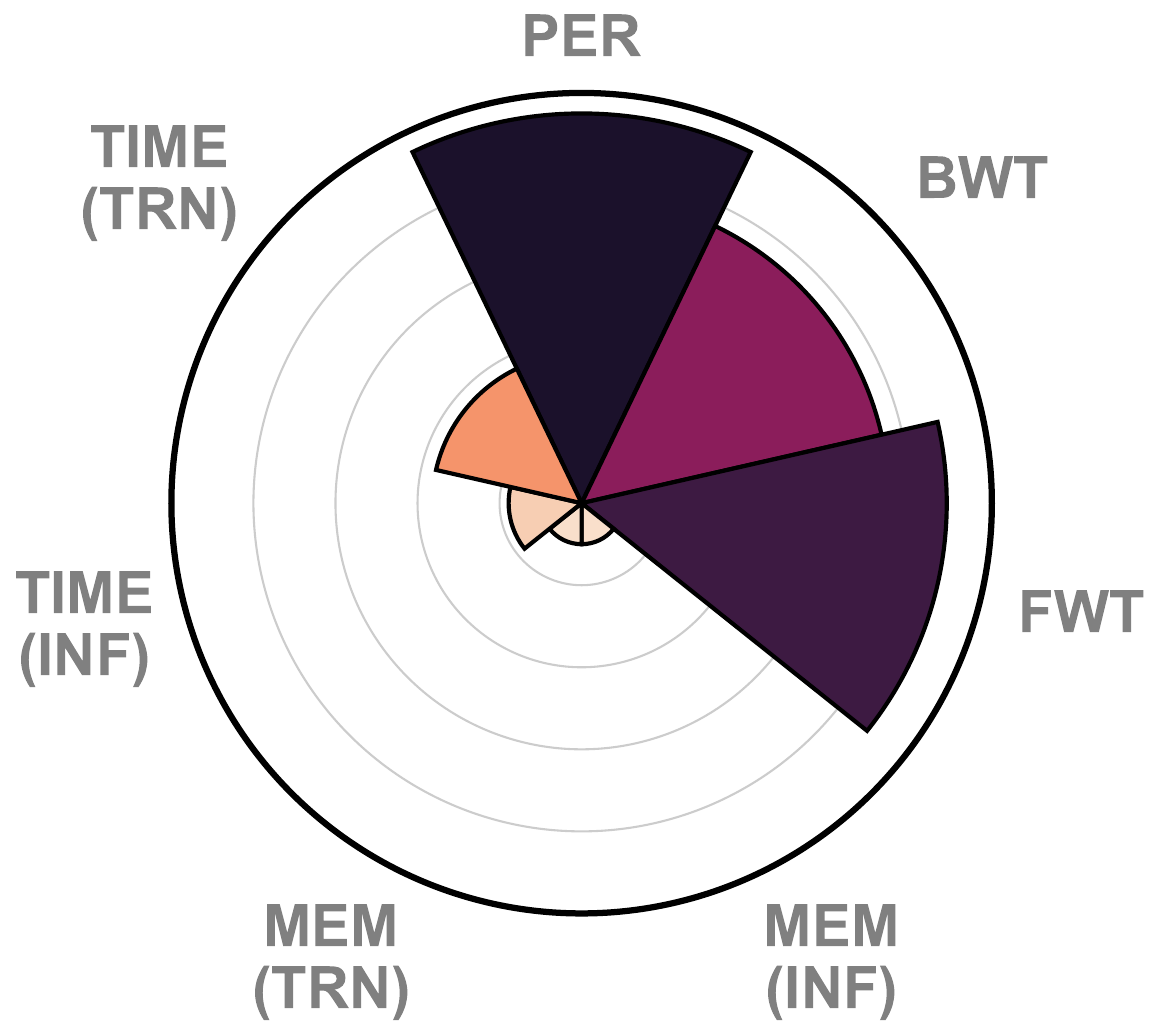}
        \subfloat{\textbf{PNN}}
        \vspace{0.25em}
        \label{fig:bench:pnn}
    \end{minipage}
    \hfill
    \begin{minipage}[h!]{00.19\linewidth}
        \centering
        \includegraphics[width=1\linewidth]{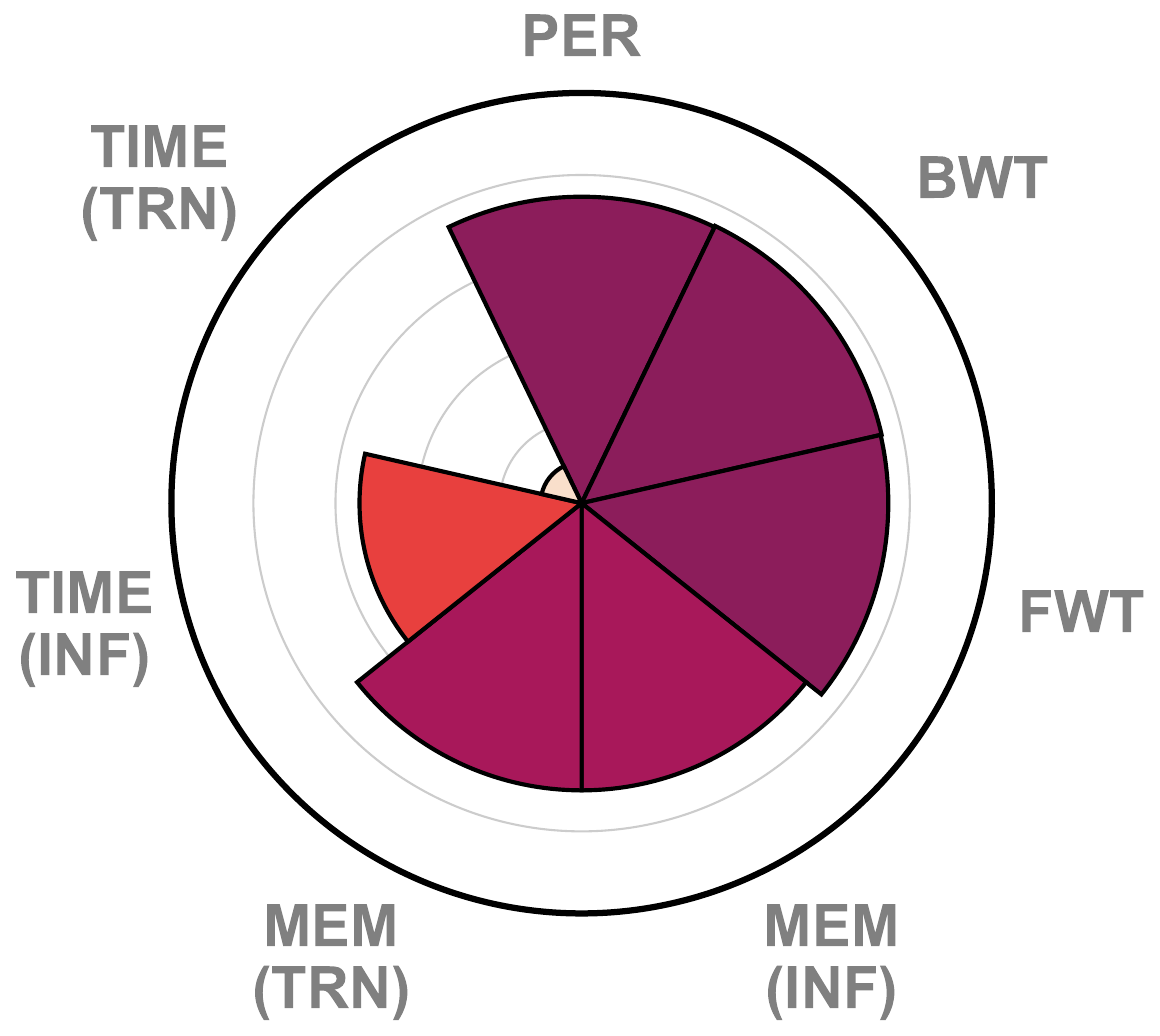}
        \subfloat{\textbf{HiSPO}}
        \vspace{0.25em}
        \label{fig:bench:hispo}
    \end{minipage}
\caption{\small
\textbf{Radar Plots of all Metrics for each CRL Method considered.}
Each sub-figure shows a \emph{radar chart} summarizing the measured metrics in Section~\ref{sec:benchmark}, as defined in Section~\ref{sec:preliminaries}, for each CRL method : \textbf{SC1}, \textbf{SCN} (Naive strategies), \textbf{FT1}, \textbf{FTN} (Finetuning), \textbf{FRZ} (Freeze), \textbf{RPL} (Replay-based), \textbf{EWC}, \textbf{L2} (Regularization), \textbf{PNN} (Progressive Neural Networks), and \textbf{HiSPO} (Hierarchical Subspace of Policies). 
For each metric, higher radial values indicate better performance, with color intensity reflecting the normalized score. 
This visualization highlights the trade-offs among memory usage, computational overhead, performance, and transfer capabilities across different approaches.
}
\label{fig:radar}
\vspace{-1.25em}
\end{figure*}

\subsubsection{Benchmark Summary }

Our evaluation reveals that, overall, CRL generally outperform single-task baselines in terms of maintaining performance across sequential tasks. Metrics such as PER, BWT, and FWT indicate that approaches incorporating architectural expansion or hierarchical decomposition tend to retain and even enhance previously learned skills better.\\
\vspace{-0.75em}

PNN and HiSPO demonstrate good performances, with PNN achieving the highest PER however with a high MEM, TRN, and INF costs. In contrast, simpler approaches such as SC1 and FRZ, though computationally efficient, suffer from significant forgetting, as indicated by negative BWT. RPL, EWC and L2 offer moderate gains but may incur additional TRN overhead.\\
\vspace{-0.75em}

In summary, our benchmark illustrates that while no single method excels across all metrics, architectural approaches (PNN, HiSPO, and FTN) present promising strategies for Offline CRL in navigation tasks. These findings underscore the trade-offs between memory cost, computational demand, and learning performance, which is an essential consideration for real-world productions and research in the field.
}

%% file: 2-MAIN/TABLES/1_gcbc_vs_hgcbc.tex
\begin{table}[h]

    \centering
    \footnotesize
    
    \begin{tblr}{
      colspec={|c||c|c||c|c|},
      row{1} = {Melon},
    }

    \hline
    
    \textbf{Task}
    & \SetCell[c=2]{c} \textbf{Success Rate} &
    & \SetCell[c=2]{c} \textbf{Episode Length} \\
    
    \hline
    
    \textbf{--} & {GCBC} & {HGCBC} & {GCBC} & {HGCBC} \\

    \hline
    \hline
    
    {S -- BASE}
        & \textbf{98.9 {\scriptsize $\pm$ 0.8}} & 97.6 {\scriptsize $\pm$ 1.7}
        & 50.9 {\scriptsize $\pm$ 2.3} & \textbf{50.8 {\scriptsize $\pm$ 2.2}} \\
    {S -- OOO}
        & \textbf{97.6 {\scriptsize $\pm$ 1.2}} & 97.1 {\scriptsize $\pm$ 2.1}
        & 55.8 {\scriptsize $\pm$ 1.4} & \textbf{55.0 {\scriptsize $\pm$ 1.7}} \\
    {S -- OOX}
        & 95.5 {\scriptsize $\pm$ 1.5} & \textbf{96.6 {\scriptsize $\pm$ 2.0}}
        & 59.4 {\scriptsize $\pm$ 1.6} & \textbf{57.4 {\scriptsize $\pm$ 1.8}} \\
    {S -- OXO}
        & 91.3 {\scriptsize $\pm$ 3.0} & \textbf{97.5 {\scriptsize $\pm$ 1.3}}
        & 60.9 {\scriptsize $\pm$ 2.5} & \textbf{55.8 {\scriptsize $\pm$ 1.8}} \\
    {S -- XOO}
        & 93.8 {\scriptsize $\pm$ 0.9} & \textbf{97.6 {\scriptsize $\pm$ 1.3}}
        & 59.5 {\scriptsize $\pm$ 1.4} & \textbf{56.1 {\scriptsize $\pm$ 1.2}} \\
    {S -- XXO}
        & 92.4 {\scriptsize $\pm$ 3.0} & \textbf{97.4 {\scriptsize $\pm$ 1.2}}
        & 68.8 {\scriptsize $\pm$ 1.8} & \textbf{66.1 {\scriptsize $\pm$ 1.3}} \\
    {S -- XOX}    
        & 94.2 {\scriptsize $\pm$ 2.5} & \textbf{98.5 {\scriptsize $\pm$ 0.9}}
        & 60.3 {\scriptsize $\pm$ 1.5} & \textbf{57.2 {\scriptsize $\pm$ 1.5}} \\
    {S -- OXX}    
        & 92.0 {\scriptsize $\pm$ 1.8} & \textbf{95.8 {\scriptsize $\pm$ 1.8}}
        & 67.8 {\scriptsize $\pm$ 2.0} & \textbf{65.8 {\scriptsize $\pm$ 2.6}} \\

    \hline
    \hline
    
    {A -- HOOO}
        & 80.2 {\scriptsize $\pm$ 3.2} & \textbf{91.3 {\scriptsize $\pm$ 3.3}}
        & 193.4 {\scriptsize $\pm$ 6.8} & \textbf{173.5 {\scriptsize $\pm$ 4.3}} \\
    {A -- HOOX}
        & 73.2 {\scriptsize $\pm$ 7.5} & \textbf{78.1 {\scriptsize $\pm$ 5.2}}
        & 222.3 {\scriptsize $\pm$ 7.8} & \textbf{215.0 {\scriptsize $\pm$ 9.7}} \\
    {A -- HXOO}
        & 92.7 {\scriptsize $\pm$ 2.1} & \textbf{97.9 {\scriptsize $\pm$ 2.3}}
        & 185.7 {\scriptsize $\pm$ 5.3} & \textbf{181.4 {\scriptsize $\pm$ 5.0}} \\
    {A -- HXOX}
        & 79.4 {\scriptsize $\pm$ 2.8} & \textbf{85.4 {\scriptsize $\pm$ 3.4}}
        & 239.4 {\scriptsize $\pm$ 2.0} & \textbf{234.2 {\scriptsize $\pm$ 3.1}} \\
    {A -- LOOO}
        & \textbf{92.7 {\scriptsize $\pm$ 3.8}} & 88.5 {\scriptsize $\pm$ 3.0}
        & \textbf{138.0 {\scriptsize $\pm$ 5.0}} & 142.0 {\scriptsize $\pm$ 4.5} \\
    {A -- LOOX}
        & \textbf{85.5 {\scriptsize $\pm$ 7.9}} & 82.7 {\scriptsize $\pm$ 5.7}
        & \textbf{171.8 {\scriptsize $\pm$ 12.6}} & 173.0 {\scriptsize $\pm$ 9.7} \\
    {A -- LXOO}
        & 89.2 {\scriptsize $\pm$ 3.5} & \textbf{90.1 {\scriptsize $\pm$ 3.8}}
        & 145.2 {\scriptsize $\pm$ 6.0} & \textbf{141.3 {\scriptsize $\pm$ 7.2}} \\
    {A -- LXOX}
        & \textbf{93.4 {\scriptsize $\pm$ 2.2}} & \textbf{93.4 {\scriptsize $\pm$ 2.2}}
        & \textbf{160.4 {\scriptsize $\pm$ 3.8}} & 161.8 {\scriptsize $\pm$ 3.9} \\
    
    \hline
    
    \end{tblr}
    
    \vspace{0.15em}
    
    \caption{
    \textbf{Performance of GCBC and HGCBC (10 seeds).} 
    HGCBC consistently outperforms GCBC in both success rate and episode length across most mazes. In some of the SimpleTown ones, the differences between HGCBC and GCBC are negligible, as these tasks are easier to learn and provide limited room for improvement with a hierarchical approach.
    }
    \label{fig:bench:gcbc_vs_hgcbc}

\end{table}

%% file: 2-MAIN/TABLES/2a_performance.tex
\begin{table}[h]

    \centering
    \footnotesize
    
    \begin{tblr}{
      colspec={|c||c|c||c|c||c|c|},
      row{1} = {Melon},
    }

    \hline
    
    \textbf{CRL Methods}
    & \textbf{AR1} & \textbf{AR2}
    & \textbf{AT1} & \textbf{AT2}
    & \textbf{ST1} & \textbf{ST2} \\

    \hline
    \hline
    
    {SC1} & 75.8 & 32.4 & 46.8 & 74.4 & 90.6 & 74.8 \\
    {SCN} & 84.3 & 84.0 & 77.2 & 87.8 & 96.9 & 96.2 \\
    {FT1} & 72.1 & 35.3 & 56.3 & 85.2 & 94.1 & 81.2 \\
    {FTN} & 83.8 & 86.9 & 81.3 & 91.0 & 97.5 & 96.8 \\
    {FRZ} & 33.4 & 65.1 & 46.8 & 70.1 & 84.6 & 64.1 \\

    \hline
    \hline
    
    {RPL} & 86.7 & 75.5 & 61.2 & 90.2 & 97.8 & 91.8 \\

    \hline
    \hline
    
    {EWC} & 72.4 & 39.6 & 56.5 & 89.6 & 94.8 & 83.2 \\
    {L2} & 72.0 & 20.3 & 48.0 & 82.6 & 93.9 & 81.1 \\

    \hline
    \hline
    
    {PNN} & \textbf{90.1} & \textbf{94.4} & \textbf{85.0} & \textbf{93.4} & \textbf{98.8} & \textbf{98.1} \\
    {HiSPO} & 85.7 & 88.3 & 75.2 & 83.5 & 97.0 & 97.1 \\
    
    \hline
    
    \end{tblr}
    
    \vspace{0.15em}
    
    \caption{
    \textbf{Performance of the CRL Methods (3 seeds).} 
    PNN attains the highest scores overall, reflecting its strong ability to retain and reuse knowledge. 
    In simpler settings, methods like FTN can achieve competitive performance, but they lag behind more adaptive approaches (RPL or HiSPO) in complex tasks.
    }
    \label{fig:bench:perf}

\end{table}

%% file: 2-MAIN/4-discussion.tex
\vspace{-0.1em}
\section{Discussion \& Future Work}
\label{sec:discussion}
\vspace{-0.1em}

{
\setlength{\parindent}{0pt}

We introduced \textbf{Continual NavBench}, a new benchmark designed for \emph{Continual Offline Reinforcement Learning} in video game–inspired navigation tasks. Our experiments show that while approaches like \textit{PNN} and \textit{HiSPO} are good at preserving and reusing skills across tasks, they may incur either significant memory or computational costs. In contrast, single‐model or regularized methods (\textit{SC1}, \textit{FRZ}, \textit{EWC}, \textit{L2}) remain lighter in resource consumption but often exhibit higher forgetting. These findings highlight the inherent trade‐offs between scalability, performance, and efficiency in Offline CRL.

Looking ahead, we believe there are several promising paths for future research. First, \emph{enlarging the environment suite} (e.g., multi‐level mazes, dynamic obstacles, or different agents) could further stress the agent’s ability to adapt in complex scenarios. Second, \emph{exploring hybrid approaches} (e.g., combining subspace methods with replay or regularization) may yield more balanced solutions that preserve prior knowledge without increasing memory or training costs. Third, incorporating \emph{domain transfer}, from offline learning to online finetuning, to shed light on how efficiently each method generalizes beyond using datasets.\\
\vspace{-0.7em}

Ultimately, we hope \textbf{Continual NavBench} spurs more rigorous and reproducible evaluations of continual learning strategies for navigation and beyond. By offering diverse tasks, open‐source tools, and well‐defined metrics, our benchmark lowers the barrier for future work to compare methods, refine architectures, and propose novel solutions. We believe that systematically addressing memory, training, and inference constraints will be pivotal in bringing \emph{lifelong autonomy} to both realistic research prototypes and real‐world production pipelines.

\vspace{0.8em}
\section{Acknowledgement}
\label{sec:thanks}
\vspace{0.3em}

This project was provided with computer and storage resources by GENCI at IDRIS thanks to the grant 2024-AD011015210 on the supercomputer Jean Zay's CSL partition.

}